\documentclass[10pt,journal,compsoc]{IEEEtran}
\ifCLASSOPTIONcompsoc
  \usepackage[nocompress]{cite}
\else
  \usepackage{cite}
\fi
\ifCLASSINFOpdf
\else
\fi

\usepackage[utf8]{inputenc} 
\usepackage[T1]{fontenc}    
\usepackage{booktabs}       
\usepackage{amsfonts}       
\usepackage{nicefrac}       
\usepackage{microtype}      
\usepackage{graphicx}
\usepackage{multirow}
\usepackage{amsmath}
\usepackage{amssymb}

\usepackage[pagebackref,breaklinks,colorlinks]{hyperref}

\usepackage[capitalize]{cleveref}
\crefname{section}{Sec.}{Secs.}
\Crefname{section}{Section}{Sections}
\Crefname{table}{Table}{Tables}
\crefname{table}{Tab.}{Tabs.}

\usepackage{url}            
\usepackage{amsfonts}       
\usepackage{nicefrac}       
\usepackage{microtype}      
\usepackage{xcolor}         
\usepackage{epsfig}
\usepackage{graphicx}
\usepackage{array}
\newcommand{\expee}[2]{{{{{\mathbb{E}}}}_{{#2}}{{\left[{{#1}}\right]}}}}
\newcommand{\expect}[1]{{{{{\mathbb{E}}}}}{{\left[{{#1}}\right]}}}

\newcommand{\gennorm}[1]{{\mid\!\mid\!{#1}\!\mid\!\mid}}
\newcommand{\frobnorm}[1]{{\mid\!\mid\!{#1}\!\mid\!\mid_{F}}}
\newcommand{\abs}[1]{{\mbox{abs}{{#1}}}}
\newcommand{\pnorm}[1]{{{\mid\!\mid\!{#1}\!\mid\!\mid}_{p}}}

\newcommand{\lonenorm}[1]{{\mid\!\mid\!{#1}\!\mid\!\mid_1}}

\newcommand{\dafsup}[1]{{\begin{array}{c}\mbox{sup}\\ {#1}\end{array}}}
\newcommand{\dafmin}[1]{{\begin{array}{c}\mbox{min}\\ {#1}\end{array}}}
\newcommand{\dafinf}[1]{{\begin{array}{c}\mbox{inf}\\ {#1}\end{array}}}

\newcommand{\ltwonorm}[1]{{\mid\!\mid\!{#1}\!\mid\!\mid_2}}
\newcommand{\linfnorm}[1]{{\mid\!\mid\!{#1}\!\mid\!\mid_\infty}}
\newcommand{\ltwonormsq}[1]{{\mid\!\mid\!{#1}\!\mid\!\mid^2_2}}
\newcommand{\vect}[1]{{{\bf #1}}}

\newcommand{\matx}[1]{{{\cal #1}}}
\newcommand{\msqrt}[1]{{{(#1)}^{(1/2)}}}
\newcommand{\trace}[1]{\mbox{Trace}{{{(#1)}}}}
\newcommand{\fidinf}[0]{{\mbox{FID}_\infty}}
\newcommand{\fid}[0]{{\mbox{FID}}}

\newcommand{\lfid}[1]{{\mbox{FID}_{l, {#1}}}}
\newcommand{\argmin}[1]{{\begin{array}{c}\mbox{argmin}\\ {#1}}}
\newcommand{\dafomit}[1]{{}}
\usepackage[pagebackref,breaklinks,colorlinks]{hyperref}

\usepackage[capitalize]{cleveref}
\crefname{section}{Sec.}{Secs.}
\Crefname{section}{Section}{Sections}
\Crefname{table}{Table}{Tables}
\crefname{table}{Tab.}{Tabs.}
\begin{document}

\title{SIRfyN: Single Image Relighting from your Neighbors}
\author{D.A. Forsyth,
\and
Anand Bhattad,
\and
Pranav Asthana,
\and
Yuanyi Zhong,
\and
Yuxiong Wang.\\
UIUC, Urbana, Illinois
}
\maketitle

\begin{abstract}
We show how to relight a scene, depicted in a single image, such that (a) the overall shading has changed
and (b) the resulting image looks like a natural image of that scene.  Applications for such a procedure include
generating training data and building authoring environments.   Naive methods for doing this fail.   One reason is that 
shading and albedo are quite strongly related; for example, sharp boundaries in shading tend to appear at depth
discontinuities, which usually apparent in albedo.  The same scene can be lit in different ways, and established
theory shows the different lightings form a cone (the illumination cone).
Novel theory shows that one can use similar scenes to estimate the different lightings that apply to a given scene, with bounded expected error.
Our method exploits this theory to estimate a representation of the available lighting fields in the form
of imputed generators of the illumination cone.  Our procedure does not require expensive ``inverse graphics'' datasets,
and sees no ground truth data of any kind.

Qualitative evaluation suggests the method can erase and restore soft indoor shadows, and can ``steer'' light around a scene.
We offer a summary quantitative evaluation of the method with a novel application of the FID.  An extension of the FID
allows per-generated-image evaluation.  Furthermore, we offer qualitative evaluation with a user study, and show that our
method produces images that can successfully be used for data augmentation.
  
\end{abstract}

\section{Introduction}

\begin{figure}[t]
\begin{center}
\includegraphics[width=0.9\linewidth]{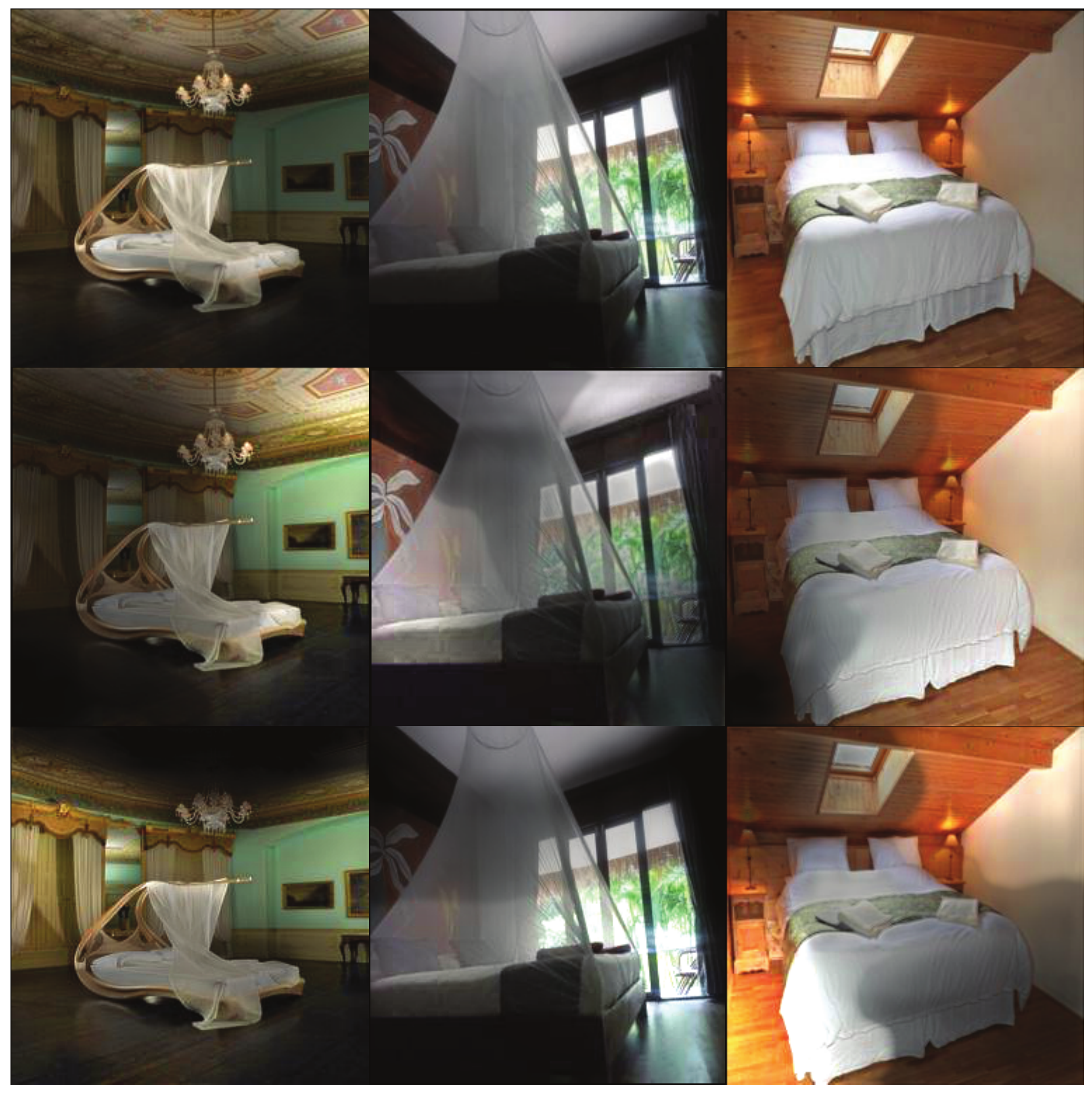}
\end{center}
\caption{Our method learns to relight images of scenes without any
  CGI data, paired data or labelled data, using two novel theorems about lighting in scenes.
  {\bf Top:} original images. {\bf Center} and {\bf bottom} rows show curated relit images of these scenes; each has the same mean
  intensity as the original image to avoid confusing effects of overall brightness.
  }
\label{fig:teaser}
\end{figure}
When someone switches on a light, the image at the back of your eye changes even when there is no other change.  The
new shading of the scene in front of you is determined by the shape and material composition of the scene, and by the
light source.  This paper shows how to {\bf relight} a scene -- we demonstrate a model that can generate images of the
scene with random (but plausible) changes in light source. One might expect scene relighting to be easy -- decompose the
scene into albedo and shading, and replace the shading with another shading field.  This approach doesn't work because
shading fields are very strongly related to albedo fields and to shape.   For example, sharp boundaries in shading
tend to appear at depth discontinuities, which are usually apparent in albedo.  This means that it is hard to produce
shading fields that apply to a particular scene with elementary methods. 
\begin{figure*}[ht!]
  \centerline{\includegraphics[width=0.85\textwidth]{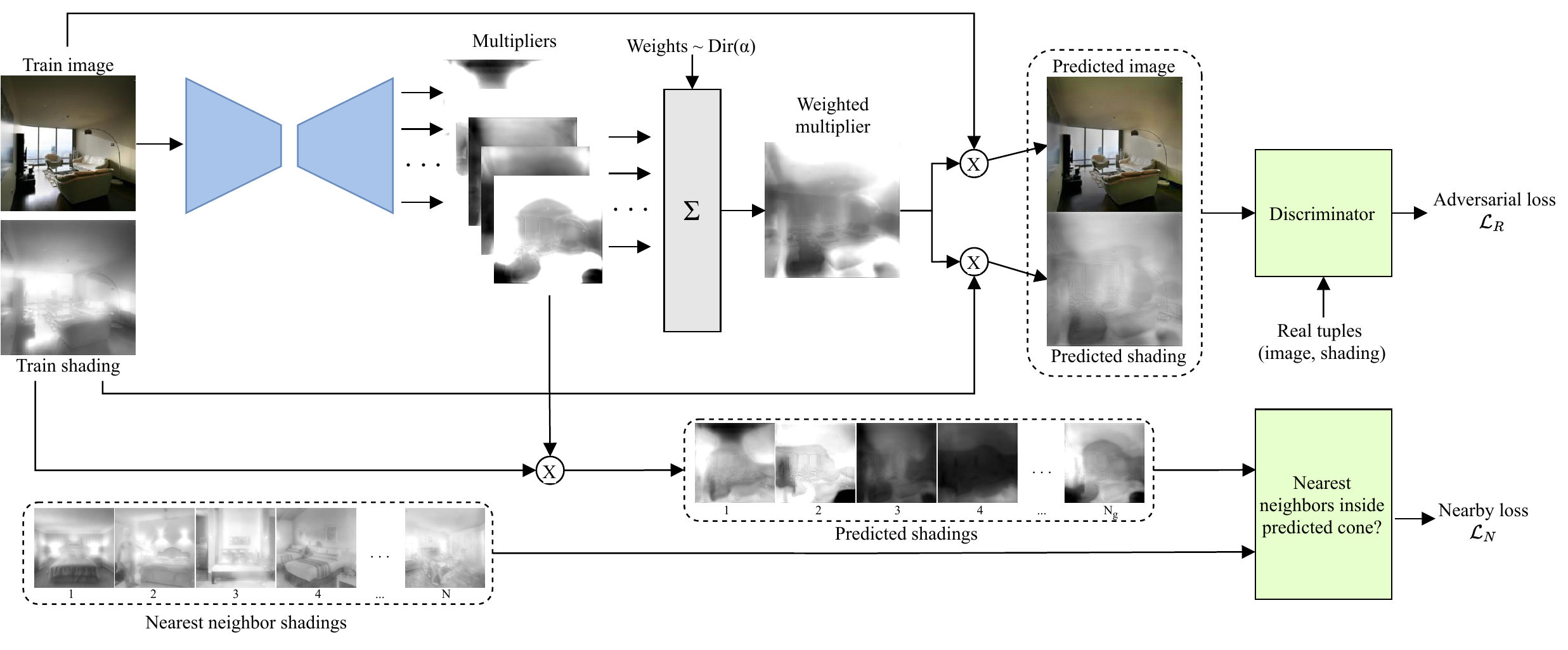}}
  \caption{\em  Our network accepts an image and produces a set of multipliers; a relighted image is produced by obtaining a random non-negative
    set of weights, then forming a weighted sum of these multipliers.   There are two main losses: the multipliers should result
    in a realistic pair of (image, shading) , imposed by an adversary; and the multipliers applied to the original image's shading field
    should encode the shading fields of nearest neighbors, a loss justified by two novel theorems \protect Section~\ref{theory}; 
\label{coverfigure} 
  }
\end{figure*}

An alternative strategy is to use ``inverse graphics'' -- learn to regress geometry,
material, etc. against an image using training set that consists of ground truth geometry, material etc. together with
computer generated images (CGI). While this strategy may offer an approach to building some kinds of vision system, it
is laborious (one must produce the training data) and is not yet reliable.  Furthermore, the inverse graphics strategy
is intellectually unsatisfying, because it doesn't explain how a visual agent learns to interpret the rich visual world
they see if the agent doesn't have a CGI training set conveniently to hand.

Instead, we see scene relighting as an unsupervised problem.   We train a neural network to produce a set of per-pixel scalings
for an image.  Each scaling produces a new image under some, likely extreme, illuminant (and, implicitly, a new shading field).  A non-negative linear combination of the scalings produces other relighting.  Two novel theorems establish that it is possible to estimate these likely relightings from observations of other, similar, scenes.
These theorems motivate using a loss that measures how well the predicted shading fields represent shading fields from other, similar scenes.
Extensive evaluation shows that (a) our relighted images are diverse and realistic; (b) rejection sampling methods are available
that can produce sharp improvements in quality; (c) our predicted shading fields are consistent with observed shading fields;
and (d) our strongest method can be used successfully as a data augmentation. 

{\bf Contributions:} We offer the first scene relighting method that requires no data marked up in any form. We offer a novel body of theory
showing how perturbations of geometry, albedo and luminaire (a luminaire is an object that emits, rather than simply reflects, light) affect scene radiosity, and that lightings of similar scenes can be used to infer lighting of
a given scene.  We show how to evaluate our models quantitatively as well as qualitatively.  We introduce a variant of the FID that applies to image
transformations and gives a per-image estimate of quality.  

\section{Related Work}

{\bf Intrinsic image decomposition} Decomposition into albedo and shading is well established; evaluation
is by comparing predicted lightness differences to human labelled versions, the WHDR score.  We use
the method of~\cite{DAFTPAMI} (see that paper for a detailed review of an area whose history dates to the mid 19'th
century),  which requires no labelled training data or CGI and produces a WHDR of 16.9\%, current SOTA for such a method.  

\dafomit{{\bf Equivariance:}  A function $\phi: {\bf x} \in X \rightarrow {\bf y} \in Y$ is equivariant under the action of a group $G$ if there
are actions of $G$ on $X$ and $Y$ such that $\phi(g \circ {\bf x})=g \circ \phi({\bf x})$. Information being
gained or lost at the boundary is an obstacle to applying the theory of group actions exactly (except for certain finite
groups~\cite{CohenWelling}).   If one relaxes the definition to require only an approximate match, well-known visual
feature representations tend to have strong equivariance properties either by design or in practice~\cite{LencVedaldi}.
Imposing equivariance properties on intrinsic image decompositions (albedo at a point should not depend on image crop)
shows sharp improvements \cite{DAFTPAMI}.  We use that paper's crude averaging strategy because we know no better.}

{\bf Inverse rendering} seeks to recover all required data for rendering from an image. 
Reconstruction methods applied to multiple images are one approach (eg~\cite{kim2016multi}); another is to impose strong
parametric models~\cite{barron2014shape}. Alternatively, one could train a regressor using  a fixed object class
(faces~\cite{tewari2017mofa}; furniture~\cite{tulsiani2017multi}; birds~\cite{kanazawa2018learning}), CGI of general shapes~\cite{li2018learning,LiSRSC20},
or multiple images~\cite{yu2019inverserendernet}.   Very strong reconstructions can be obtained from rendered images~\cite{li2018learningspat};
isolated object reconstructions can be relit~\cite{DBLP:conf/cvpr/ZhangLWBS21}.  Inverse rendering recovers strong scene models for indoor
scenes from CGI~\cite{DBLP:journals/corr/abs-2109-06061}, but we are aware of no successful attempt to change luminaires in images of
scenes by relighting an inverse rendering result.  In contrast to inverse rendering methods, we use no CGI and no marked up data.

{\bf Precomputed radiance transfer} (PRT)
is a method from computer graphics that builds a relightable representation of a scene by rendering
versions with multiple distinct illuminants, producing an estimate of the light transport matrix mapping
illuminant representation to relit scene  (originally~\cite{prtorigin}; review in~\cite{Ramamoorthi:2009:PBR}; compare a PRT operator with our EGM, below).  The operator can be estimated for real scenes with
a projector and a large number of images (a generalization of~\cite{BK98}) or from a collection of outdoor images~\cite{5206753}.  There are strong regularizers which 
can massively reduce the amount of scene data needed (eg~\cite{ramaeccv,prthart}). Given relatively
few lightings of a scene, a neural network can produce a very good light transport matrix estimate~\cite{ren2015image,xu2018deep};
this is consistent with theorems 1 and 2.  In contrast to PRT methods, our estimate requires only one image of a given scene, but must be a great deal less accurate.

{\bf Image relighting} is now an established task.  For scenes,
there are workshop tracks (eg~\cite{ntire,aim}),
challenges~\cite{DBLP:conf/eccv/HelouZSTABXCLWL20,DBLP:journals/corr/abs-2104-13365} and datasets
~\cite{elhelou2020vidit,murmann19}. Existing work learns image mappings
(pure image mappings as in~\cite{DBLP:journals/corr/abs-2006-07816}; depth guided, as in~\cite{9523057};
using wavelets, as in~\cite{DBLP:conf/eccv/PuthusserySKV20}; shadow priors, as in~\cite{DBLP:conf/eccv/WangSLLL20}).  In all cases,
methods are learned with paired data (ie images of the same scene under different illuminations), available in the VIDIT dataset~\cite{elhelou2020vidit} and the MIE dataset~\cite{murmann19}. VIDIT data is CGI, and emphasizes point light sources with strong shadows, which are uncommon in
indoor scenes.   Pairing is necessary to ensure that their method preserves scene characteristics~\cite{DBLP:journals/corr/abs-2110-12914,9423347}. Chogovadze {\em et al} use a multi image dataset and lightprobes to learn a scene relighter, and show the resulting augmentations improve two patch matching tasks~\cite{chogovadze2021controllable}. Philip {\em et al.} learn geometry aware models to relight outdoor video from renderings under multiple light conditions~\cite{DBLP:journals/tog/PhilipGZED19}.  Similarly, face image relighting
requires multi-illumination data (from real light stages~\cite{DBLP:journals/corr/abs-1905-00824}; made using deformable models~\cite{DBLP:conf/iccv/ZhouHSJ19}). In contrast to these methods, we do not use paired data.

{\bf Shadow synthesis} is now an established task.  People in video can be used as scene probes to learn
to create shadows for objects inserted in outdoor scenes~\cite{Wang2020PeopleAS};  methods with sufficient training data
can learn to create soft attached shadows for objects that have been inserted into indoor scenes~\cite{Sheng_2021_CVPR,Zhang2019ShadowGANSS}. In contrast to these methods, we use no paired or synthesized data, and we create entire lighting fields.

{\bf Global illumination bounds:} Section~\ref{theory} establishes that scenes with similar albedo, luminaire  and
geometry have similar radiosity.  Bounds on the effect of luminaire difference are well known
(eg \cite{Arvo}, sec 5.3; \cite{dutre}).  Previous results on the effects of albedo changes are rare (~\cite{Koenderink} sketch the case for
an integrating sphere;~\cite{Arvothesis}, p175, has a bound). There are few results on the effect of perturbing
geometry;~\cite{JacKupp} estimate the effect of flat ports on an integrating sphere;~\cite{Koenderink} argue that
sufficiently small deformations of an integrating sphere do not affect its function; and~\cite{Arvothesis}, p176, says
such results are hard to get.   

{\bf The illumination cone and its generators (ICGs):}  The family of images created by lighting a single convex diffuse object,
viewed in a fixed camera, with a point source is naturally a convex cone (reflection is linear; 
irradiance is non-negative). As Kriegman and Belhumeur~\cite{BK98} point out, this fact extends to any diffuse scene.   If
the observed scene has only a discrete set of normals -- for example, a polyhedral object -- then the cone is polyhedral
and has a finite set of generators (i.e. any element of the cone is a non-negative sum of a finite set of
generators). In simple cases, one can recover generators by illuminating the object with different sources.
Interreflection and shadowing may mean that the cone is not polyhedral, but experiments suggest that (a) relatively few
generators provide a good representation of the cone anyhow (\cite{BK98}, p.8;~\cite{GBK01}, p.644) (b) the cone is relatively
``flat'' (\cite{BK98}, conjecture 1; p.12; for convex objects~\cite{BasriJacobs}) and (c) illumination cone methods offer strong models of extreme relighting
for face images \cite{GBK01}.  

\dafomit{{\bf Object insertion} starts with Lalonde \emph{et. al.}~\cite{lalonde2007photoclipart}, who insert fragments into target
images.  Karsch et al show that computer graphics objects can be convincingly inserted into inverse rendering models
obtained with geometric inference~\cite{karsch2011rendering} or with single image depth
reconstruction~\cite{karsch2014automatic}. Inverse rendering trained with rendered images can produce excellent
reshading of computer graphics objects~\cite{ramachandran1988perceiving}.  However, recovering a renderable model from
an image fragment is extremely difficult, particularly if the fragment has odd surface texture.   Liao {\em et al.}
showed that a weak geometric model of the fragment can be  sufficient to correct, if one has strong geometric
information about the target scene ~\cite{liao2015cvpr,liao2019IJCV}.   In contrast, our  work is entirely image based: 
we take a scene, and relight it.}

\section{SIRfyN: Methods and Network}
\label{predicting}

At run time, we want to be able to produce multiple realistic relightings from a single image of a scene, without requiring
intrinsic image decomposition etc.    We will build a network that maps a 256x256x3 color image $I_V$ of a scene $V$ to
a set of $N_g$ multipliers $M_i(I_V)$  (256x256x1 scalar fields).  For non-negative $\theta_i$ and $\cdot$ broadcast element-wise multiplication,
these multipliers will have the property that  $M(\theta) \cdot I_V=\left[\sum_i M_{i}(I_V) \theta_i\right]\cdot I_V$ is a relighted version of $I_V$.
 One can think of these multipliers in terms of shading fields, by modelling
the image as a product of albedo and shading $I_V=A_V \cdot S_V$; then relighting $I_V$ by $M(\theta)$ implies a new
shading field $S_V \left[\sum_i M_{i, V} \theta_i\right]$.  So predicting $M_i(I_V)$ from $I_V$ yields an encoding of
the shading fields we expect for that scene -- equivalently, estimates of generators for the scene's illumination cone.

 {\bf Network:} We predict multipliers using a network to obtain $\left[M_1(I), \ldots,   M_{N_g}(I)\right]=N(I;
 \phi)$. We implement $N(I; \phi)$ using a u-net, with skip connections (details in supplementary; $\phi$ 
parameters). The output layer forms a multiplier directly from activations, with housekeeping losses to ensure that multipliers are non-negative.

The key trick is obtaining losses to train such a network.  Section~\ref{theory} sketches two novel theorems we rely on
(proved in detail in supplementary).  Theorem 1 states that, for two scenes $V$ and $V'$ with similar
geometry, illumination and albedo, the difference between radiosity $B_V$ and $B_V'$ is bounded (despite the effects of
interreflections).   Now radiosity is linear in illumination and images are linear in radiosity, so we can model
different relighted images of the same scene with some linear operator, which is easily estimated if we have multiple
images of the same scene under different lights.  We don't have such images, but Theorem 2 states that the expected future error of
using images of similar scenes in place of images of the original scene is bounded.

We can now test whether the network is producing sensible multipliers with theorem 2.
Theorem 2 means that, for the shading field $S_V'$ of each  image of a similar scene $I_V'$, there should be
some non-negative $\theta_{V'}$ so that $S_V' \approx S_V \left[\sum_i  M_{i}(I_V) \theta_{V', i}\right]$.

{\bf Finding images of similar scenes:}  It is not practical to enforce the exact constraints of section~\ref{theory}, but the intention of these
constraints are that a nearby image should depict a scene of similar shape to the original scene.  GIST features were hand-tuned
to encode the overall shape of a scene~\cite{Oliva:2001wt}, and we use the 20 nearest neighbors in GIST features.  We have not attempted to optimize further,
but informal experiments with learned embeddings were not successful, and we expect that incorporating an albedo
matching term might help.

{\bf Exploiting images of similar scenes:}  Assume we have $k$ images which can be used to estimate generators
for an image $I_V$. There should be some non-negative $\vect{w}$ such  
that the shading field $S_i$ of the $i$'th such image is encoded by the multipliers, so $S_i  \approx S_V \left[\sum_i M_{j} w_j\right]$.
Geometrically, we require that each $S_i$ lie inside the cone generated by $\left[S_V M_1, \ldots, S_V M_{N_g}\right]$.
There are some subtleties involved in deriving a loss from this constraint.  
We do not want $2 S_i$ to be ``further outside'' the cone than $S_i$ is,  so
the loss needs to be independent of the scale of the shading field. We achieve this by rescaling all shading fields to have mean
0.7 (the value is arbitrary, and affects only the scale of the loss).  The nearby loss on multipliers 
can be written as
\[
{\cal L}_N(I_V,
\phi)=\sum_{i \in {\cal N}} C(S_i, M(I_v))
\]
for ${\cal N}$ the neighbors of $I_V$ and
\[
C(S_i, M(I_V))=\begin{array}{c} \mbox{min}\\ \vect{w}| \vect{w}>0\end{array} \ltwonorm{S_i-\sum_j
      S_V M_{j}(I_v; \phi) w_j}^2
\]
but this is difficult to evaluate (the non-negativity constraint means the value is a solution to a quadratic program).
We estimate  $C(S_i, M(I_V))$  by solving for $\vect{w}$ using least squares; clipping to ensure $\vect{w}\geq 0$; then taking a $N_{gd}$ steps of projected gradient descent in $\theta$.  This yields a usable
approximate loss.  

{\bf Realistic scene preserving shading:}  It is important that, for predicted multiplier
$M(\theta)$, the image $M(\theta) I_V$ should look like a relit version of $I_v$.  We impose this constraint using an adversarial
loss to adjust the joint distribution of (predicted shading, predicted image) to match the
best estimate of the true distribution.  Here  ${\cal L}_{R}(I_V; \phi)$ is the value of an adversary that compares generated tuples
$(M(\theta)S_V,M(\theta)I_V)$ to real tuples $(S_I, I)$ and $S_I$ is the shading field of image $I$, 
estimated using the method of~\cite{DAFTPAMI}.    The discriminator score (we use a hinge loss, as in~\cite{lim2017geometric};
details in supplementary) depends only on local neighborhoods, as in PatchGAN~\cite{Isola_2017_CVPR}, and
the size of the neighborhood has important effects on results.  

{\bf Extrapolation with a barrier:} If nearby images for a given image have relatively similar shading fields, the nearby loss will not encourage diversity because it does not
force the $M_{V, j}$ to be different.  But we want to extrapolate and force the new shadings to be as extreme as
possible.  Write $\mbox{int}$ for the operator that computes the intensity of an image,
$\overline{I}$ for the
pixelwise mean of an image,
$P_V=\mbox{int}(M_V I_V)$ and $Q_V=\mbox{int}(I_V)$. We use
\[
  {\cal L}_{B}(I_V; \phi)=\overline{L} \mbox{ where } L=-\log \mid\!\frac{P_V}{\overline{P_V}}-\frac{Q_V}{\overline{Q_V}}\!\mid
  \]
  as a diversity loss; this forces multipliers to be different from 1, but may result in unrealistic relightings if
  overweighted.
  
\dafomit{
    }
{\bf Realism housekeeping:} we require that any pixel in $R$ should be non-negative (${\cal L}_U(R)$), and less than one
(${\cal L}_O(R)$), and achieve this with simple one-sided quadratic losses.  Note that this constraint applies to
relighted images, rather than shading fields -- applying the constraint to shading fields will lead to regions with dark
albedo being consistently dark.  We show the network $A$ as well as $I$ so that it can anticipate this loss and produce
very bright shading at locations where the albedo is dark.

{\bf Overall loss:} the loss is then:
\[
  {\cal L}={\cal L}_N +  \lambda_B {\cal L}_B+\lambda_O {\cal L}_O + \lambda_R {\cal L}_R +\lambda_U {\cal L}_U
  \]

\section{Theory: ICGs and EGMs}
\label{theory}

{\bf Light in diffuse scenes:} The brightness of  each pixel in the camera is determined by the radiosity at the corresponding scene point.  
Write $\rho(\vect{u})$ for albedo at the point on the geometry parametrized by $\vect{u}$, $E(\vect{u})$ for the
diffuse emittance of that point (only non-zero when the point is actively emitting light, so  a point on a luminaire with non-zero weight), and $B(\vect{u})$ for the
radiosity at $\vect{u}$. Write ${\bf D}_\rho$ for the linear operator that maps $f(\vect{x})$ to
$(\rho f)(\vect{x})$. Then standard diffuse interreflection theory (see,
eg~\cite{nn36019,nn36017,SillionBook,CohenBook}) gives
$B=E+ {\bf D}_\rho {\bf K}  B$, equivalently $(I-{\bf D}_\rho {\bf K}) B={\bf R} B=E$
where ${\bf K}$ is a bounded linear integral operator.  This yields a Neumann series
solution $B={\bf R}^{-1}E=E+{\bf D}_\rho {\bf K} E+ ({\bf D}_\rho {\bf K})^2 E+ \ldots$
which should be interpreted as ``emittance+direct term+one bounce+...''; for most practical geometries and albedos,
relatively few terms at relatively poor spatial resolution yield an acceptable solution (eg~\cite{rushmeierthesis}, reviews in~\cite{dutre,arikan:2005:globalillumination}).
\dafomit{
because ${\bf D}_\rho {\bf K}$ is bounded
by $p=\mbox{sup}_{\vect{x}} \rho$ and so ${\bf R}^{-1}$ is bounded by $1/(1-p)$ (eg \cite{Arvo}, sec 5.3).
This bound is relatively loose for practical geometries, and rendering practice
tends to treat the bound on ${\bf D}_\rho {\bf K}$ as a small positive number \cite{dutre}.  The bound easily yields bounds on
how radiosity changes when the illumination changes (eg~\cite{Arvo}; below).  
Previous results on the effects of albedo changes are rare (~\cite{Koenderink} sketch the case for
an integrating sphere;~\cite{Arvothesis}, p175, has a bound), but they are easy to obtain (below).  We have found no results on the effect of perturbing geometry, which appear to be hard to obtain (cf~\cite{Arvothesis}, p176;~\cite{JacKupp} estimate the
effect of flat ports on an integrating sphere).}

{\bf ICGs from diffuse interreflection theory:} Assume that a scene has $N_e$ luminaires, so that $E=\sum_i \theta_i E_i$ with
$\theta_i\geq 0$ the intensity of the luminaires.  We choose $\ltwonorm{E_i}=1$. Because the solution is linear in the emittance, we can write
$B(\vect{u};\theta)= {\bf R}^{-1}\sum_i \theta_i E_i=\sum_i \theta_i {\bf R}^{-1}E_i=\sum_i \theta_i
B_i(\vect{u})$ -- if we know the solution for each luminaire separately, we can construct a solution for
any linear combination of luminaires.  If we have one image of the scene for each luminaire, where the image is illuminated by
that luminaire alone, we can generate an image from this viewpoint for any combination of the luminaires.   But we
do not have these images.

{\bf Similar scenes with similar luminaires have similar radiosity:}
We assume that all relevant reflection is diffuse; it is likely that this can be generalized (discussion).
A scene $V$ is a tuple of geometry, albedo and luminaire model; write $V=\left({\cal G}, \rho,
{\cal E}\right)$, where ${\cal E}$ is the luminaire model.     The geometry ${\cal G}$ consists of a set
of surfaces parametrized in some way and a 2D domain for that parametrization; write ${\cal G}=(\vect{s}(\vect{x}), {\cal
  D})$.  The luminaire model captures the idea that lighting in a particular scene may change.  We have
${\cal E}=\left(E_1, \ldots E_{N_e},  P(\theta)\right)$ for a set of potentially many basis luminaires and a probability distribution over the coefficients.
This information implies a probability distribution over luminaires for the scene $P(E)$.   {\bf Similarity:} Two scenes $V$ and $V'$ are similar if:
there is some affine transformation so that
 ${\cal G}'({\cal G}, \matx{A}, \vect{b})=(\matx{A} \vect{s}(\vect{x})+\vect{b}, {\cal D})$.
$E'_i(\vect{x})=E_i(\vect{x})$ ; and $P_{V}(E)$ and $P_{V'}(E)$  the same.  Consider some lighting $E_V$ of $V$ and a different lighting $E_V'$ of $V'$:
Theorem 1 (below) establishes that if $V$ and $V'$ are similar,
and $E_V$ and $E_V'$ are similar,  then $B_{V}$ will be close to $B_{V'}$.

{\bf Theorem 1:} {\em For $V=({\cal G}, \rho, E)$ and $V'=({\cal G}'({\cal G}, \matx{A}, \vect{b}), \rho', E')$, where
  $\epsilon_E\gennorm{E} = \gennorm{E-E'}$, $\epsilon_\rho=\mbox{sup}_{\cal D}\abs{\left[\rho-\rho'\right]}$, $p=\mbox{sup}_{\cal D} \rho$,
  $p'=\mbox{sup}_{\cal D}\rho'$, $c$ is the condition number of $\matx{A}$ (ratio of largest to smallest eigenvalues), 
  we have:}
\[
  \gennorm{B_{V}-B_{V'}} \leq c_1(\epsilon_E, \epsilon_\rho, p, p', c)
\]
{\bf Proof:} Elaborate, relegated to supplementary, which gives the form of $c_1$

{\bf The effective generator matrix (EGM) of a scene:}  Choose some $N_o$ element orthonormal
basis to represent the possible radiosity functions for the scene (if $N_o=N_e$, the representation is exact,
because there is a finite dimensional space of luminaires).  In this basis, any particular radiosity $B(\vect{x})$ has
coefficient vector $\vect{b}$.  Write $\vect{b}_i$ for the coefficient vector representing
$B_i(\vect{x})$; and $\matx{B}=\left[\vect{b}_1, \ldots, \vect{b}_{N_e}\right]$.
For any $r \leq N_e$, an {\bf effective generator matrix} for scene $V$ is an $N_o \times r$ orthonormal matrix  
$\matx{M}_{V}$  such that for any illumination condition producing radiosity represented by
$\vect{b}$ , there is some $\vect{w}$ such that the radiosity  represented by $\matx{M}_{V, r} \vect{w}$
is similar to the actual radiosity field.   In particular,
we want
\[
{\cal L}(\matx{M}_{V})=\expee{\mbox{inf}_{\vect{w}}\ltwonorm{\matx{M}_{V} \vect{w}-\matx{B}\theta}^2}{\theta \sim P(\theta)}
\]
to be small; $\matx{M}_{V}$ is clearly not unique. Now if we have $k$ radiosity fields for  a scene under different,
unknown, illumination conditions, for $k$ very large, where for each $\theta_i \sim P(\theta)$ (which is not known).  We
could estimate an EGM by replacing the expectation with an average, and minimizing.  But we do not have $k$ images of the same 
scene. 

{\bf Estimating an EGM:} An EGM can be estimated from radiosities obtained from similar scenes.
Write ${\cal L}_{T}(\matx{M})$ for the true expected error of using an EGM $\matx{M}$ to represent the radiosity of
a scene.  Theorem 2 shows that substituting an estimate $\hat{\matx{M}}_O$ obtained by using $k$  radiosities in total, taken from distinct similar scenes, for the best (but unknown) effective generator matrix $\matx{M}_O$ incurs bounded error.  This means we can use estimates from similar scenes 
with confidence, if the scenes are similar enough.

{\bf Theorem 2:}{\em  ${\cal L}_{T}(\hat{\matx{M}}_O)-{\cal L}_T(\matx{M}_O)$ is bounded.}\\
{\bf Proof:} Elaborate; relegated to supplementary, which provides the bound.

{\bf Image loss from theory:}   Note first that theorem 2 applies to images as well as to scenes. If one
views a scene through a fixed camera, the camera is a linear mapping (taking scene radiances to pixel values)
and it is bounded.  The result is very powerful, because it says that an estimate of the generator matrix for a given image 
can be obtained from enough images of similar scenes --- take the shading fields $S_i$ of similar scenes and form a $\matx{M}$ that
minimizes
\[
\sum_i {\mbox{inf}_{\vect{w}}\ltwonorm{\matx{M} \vect{w}-S_i}}.
\]
Using this approach directly is inconvenient because it means we need to find similar images at inference time.
Instead, we derive a loss from Theorem 2 and use a network to predict the generator matrix.

\begin{figure*}
  \centerline{\includegraphics[width=\textwidth]{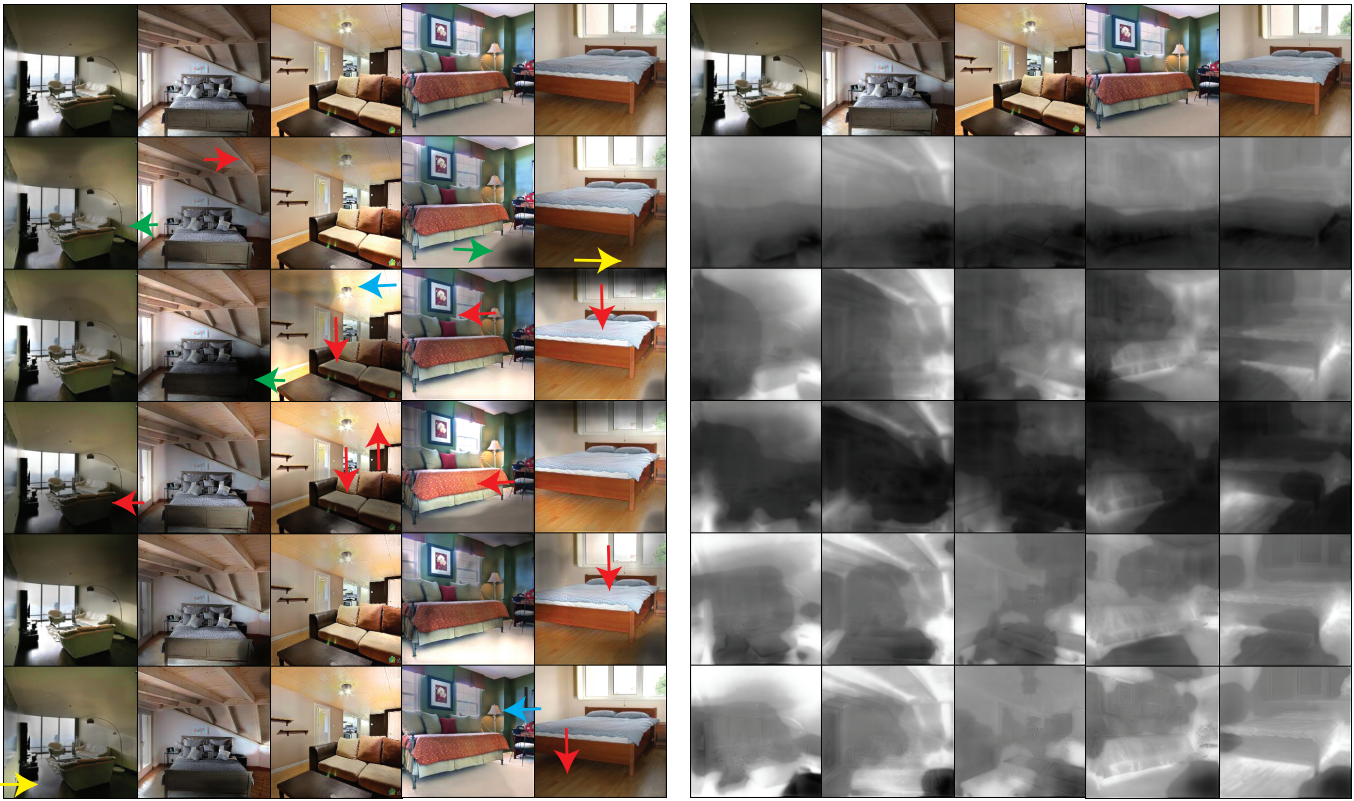}}
\caption{\em
  Images relighted at $\alpha=0.1$ (more variable; {\bf left}; cf  $\alpha=0.5$ less variable in supplementary) and the first five multiplier fields for
  each {\bf right}.  {\bf Top row:} shows original images;
  the following five rows are random relights, {\em not curated}.  A number of important effects are notable:  {\bf green} shadow removed or replaced;
  {\bf red} significant brightening or darkening of one surface, registered to geometry. {\bf yellow} suppression or enhancement of a gloss effect; {\bf blue}
  suppression or enhancement of a luminaire.
  All relights are shown at the same mean intensity as the original image, to ensure that effects are not masked by overall shift in brightness.  Multiplier fields are scaled so
  that one is mid-grey, less than one darker, greater than one brighter.  Recall relights are combinations of
  multipliers, with coefficients drawn as in Section~\ref{predicting}
  \label{examples}
  }
\end{figure*}

\section{Quantitative Evaluation for Relighting}

{\bf Realism} in current image generators is typically measured with the FID~\cite{FID}. One embeds a set of generated images, then
computes the Frech\'{e}t distance between the resulting points and an embedding of a set of real images.
We use the standard VGG embedding of~\cite{FID}, with dimension $d_f=2048$.
This is an estimate of the difference between the distributions from which the samples were taken; smaller values are
better.  As~\cite{unFID} show, the FID estimated from a finite set of points is biased by a $1/N$ term that depends on 
details of the generator, but straightforward extrapolation methods remove this bias successfully to produce $\fidinf$.
An important difference between image relighting and image generation is that relighting should map the collection of real images to
itself, preserving frequencies etc. (loosely, relighting is an {\em isomorphism}).  In turn, this means that the FID
between a set of images ${\cal O}$ and a relighted set ${\cal R}$ of images should be zero, so $\fidinf({\cal O}, {\cal
  R})$ should be zero up to sampling error.   Note strong realism scores can be obtained by not changing the original image much.

{\bf Diversity:}   We must also evaluate whether relighted versions of an
image are actually different from the original image.  We measure {\bf diversity} by comparing relighted images to the
original.  We rescale the relit image to have the same mean as the original image, so that just scaling images will get
zero diversity score.  
Write $N_t$ for the total number of pixels in relighted images; we score diversity with $MSD$, computed as
\[
  MSD_{\hat{\theta}}=\sqrt{\frac{1}{N_t}\sum_{i \in \mbox{images}}\overline{D}^2_i},
  \]
  where $D_i=(I_i-R_i (\overline{I}_i/\overline{R}_i))$.
Note quite unreasonable shadings may have high diversity scores. 

\begin{figure*}
\centerline{\includegraphics[width=\textwidth]{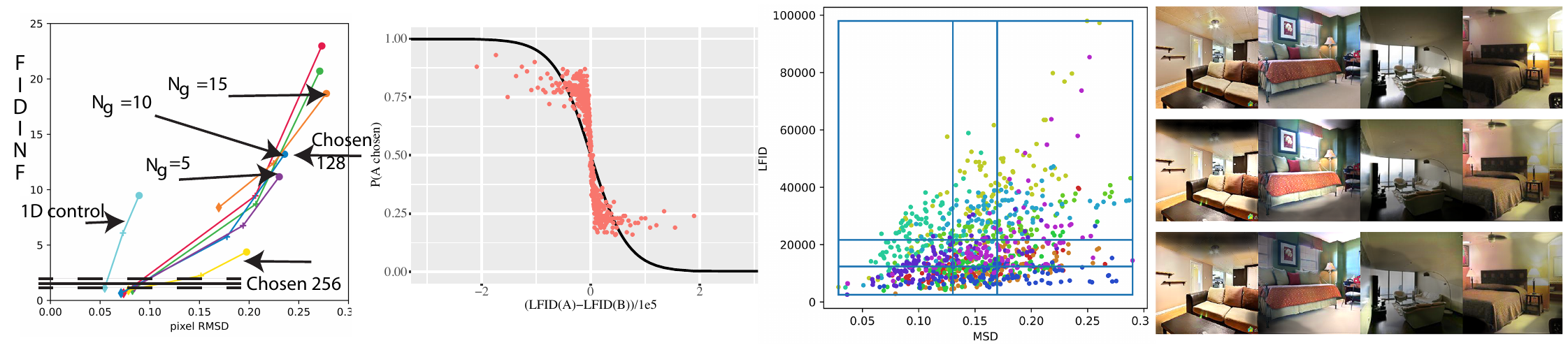}}
\caption{\em
  {\bf Left:} Model selection is by inspecting plots of $\fidinf$ against MSD  for three different values of $\alpha$ (0.01, 0.1, 3.0).  Best behavior is down and to the right (low $\fidinf$, high MSD). We  show some effects here; more in supplementary.    Models seem largely to stick fairly close to a Pareto frontier.
  The pixel uniformity loss is unhelpful   (supplementary) and so its weight $\lambda_P$ is 0 except where noted.
  There is a bias variance tradeoff in the number of generators; $N_g=15$ gives
  more diversity for weaker quality, etc.; $N_g=10$ is about right, compare \protect~\cite{BasriJacobs}.  Training at
  256x256 produces notable improvements (compare chosen  at 128 with chosen at 256).  A pure mapping of pixels (1x1
  convolutions, trained with the same losses; {\bf 1D} control) is a lot weaker than any of our models.  Horizontal line shows $\fidinf$  of random splits of data ({\bf Spl} baseline); dashed are 1-sd limits (mean=0.99, sd=0.027). 
  {\bf Center left:}  Difference in LFID is a good predictor of human subject preferences.  Points show probability that A is selected against difference value,
    for each of 500 quantiles of LFID difference and a total of 50000 preference tests.  Curve shows a logistic regression predictor; the fit would likely be
    improved by removing large differences in LFID.  Choice of scale has occluded two data points (one at very far
    right; one at very far left).
  {\bf Center Right:} LFID plotted against MSD for 100 randomly relighted samples for 10 test images.  Note the wide spread; a
  rejection sample (which seeks large MSD for low LFID, say), is likely to be successful.  {\bf Right:} Rejection results: {\bf RTop:} original images;
  {\bf RCenter:} relights with highest MSD for $LFID<25000$ (so most different from original at fixed quality);
  {\bf RBottom:} relights with lowest LFID for $MSD>0.2$ (so best at fixed difference from original).
  Best viewed in color at high resolution, compare supplementary material.
    All relights are shown at the same mean intensity as the original image, to ensure that effects are not masked by overall shift in brightness.
}
    \label{bigfig}
\end{figure*}

{\bf Local FID:} Because relighting is an isomorphism, we can recover a per-image measure of quality.  Assume we obtain ${\cal R}$ from
${\cal O}$ by relighting a single image $I_k$ to obtain $R_k$ and compute $\fid({\cal O}, {\cal R})$.  This value -- which is
non-negative, and is ideally zero -- measures the extent to which the relighted image is realistic.  Furthermore,
different relightings of the same image will have different values.  This value is the {\bf local FID} of  image $R_k$.  This
local FID strongly reflects human preferences (Figure~\ref{bigfig}; people reliably prefer images with smaller local FID).
Computing the local FID requires care; the supplementary shows how.

\begin{figure}
  \centerline{\includegraphics[width=\columnwidth]{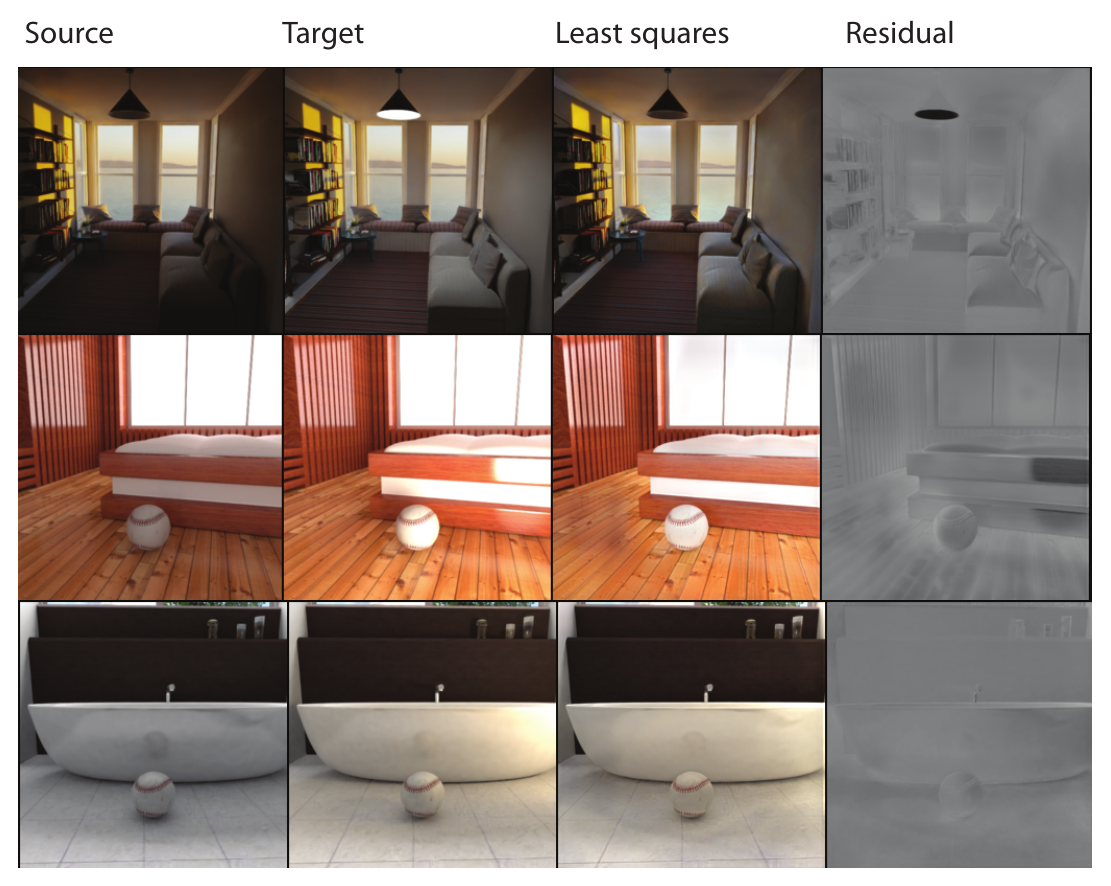}}
\caption{\em
  We investigate the plausibility of generators by: rendering CGI scenes under two lights; computing the
  generators for the {\bf source}; constructing weighted sum of generators that yields a least squares fit to the {\bf
    target}; then comparing target to the {\bf least squares}.  The {\bf residual} error in intensity
  is shown on a scale where white is 1; mid grey is 0; and black is -1.  Note relatively good fits, suggesting that
  for the cases we have investigated, generators are a reasonable fit.  {\bf Quantitative:} over a total of 11 pairs
  (22 tests), for a dataset we will provide, we find average metrics between the least squares prediction and the target of
  RMSD:   0.0503  PSNR: 26.6  LPIPS: 0.0648, suggesting quite successful prediction of the target
  by source generators.
  \label{gentoCGI}  }
\end{figure}

\section{Experimental Details}
\label{experiment}

{\bf Datasets:}  No markup is required. Training albedos are obtained with the method of~\cite{DAFTPAMI}.  
We train with images from the LSUN Bedroom dataset (\cite{yu15lsun}), resized as described.
We use 20,000 images to train the generator (augmented by crop and resize and horizontal flips)
and train the discriminator with 20,000 distinct images (to prevent shading ``leaking'').
Nearest neighbors are drawn from 100,000 LSUN Bedroom images,  and are augmented in the
same way (so when a training image is cropped, the same crop is applied to each nearest neighbor).
We use 20 nearest neighbors for each image to compute ${\cal L}_N$.

{\bf Relighting weights in training:} During training we must ensure that any non-negative combination of multipliers results in
a realistic image; we do so by randomly selecting one multiplier per image in a batch, and passing those to the adversary.

{\bf Choice of hyperparameters:} For model search purposes, we train our network at 128x128.
Simple qualitative experiments suggest the  model
is not particularly sensitive to the choice of $\lambda_R$, $\lambda_U$ and $\lambda_O$,and we have used $1, 100, 100$.
Simple qualitative experiments show that training without projected gradient descent ($N_{gd}=0$) fails reliably; more than one
step ($N_{gd}>1$) slows training without offering any obvious benefit.  Other hyperparameters that have significant effects are:
$N_g$,  $\lambda_P$, $\lambda_B$ and the scale of the discriminator.  Figure~\ref{bigfig} plots $\fidinf$ against MSD for a
variety of methods at three different values of $\alpha$ (FID vs MSD and detailed comparisons in supplementary).   In summary, we
have: a small scale discriminator is better;  $\lambda_B=5$ is about right; there is a bias-variance tradeoff for $N_{g}$, and
$N_g=10$ is about right (cf~\cite{BasriJacobs}).
\dafomit{; $\lambda_P$ should be $0$ unless one seeks extreme reshadings without regard to quality.}

{\bf Inference and equivariance:} Our chosen network is trained on 256x256 images.
The model should be sensitive to geometry (and is - supplementary), so resizing an image that isn't
square should be dangerous (and is - supplementary), and so we always work with 256x256 images, 
and crop (rather than resize) as required.   We model the weights as distributed according to a symmetric Dirichlet distribution with concentration parameter $\alpha$.  Then we generate relighted images by: drawing a sample of weights; computing multipliers; computing a weighted combination of
multipliers; and multiplying by the image.   

{\bf Baselines and Controls:} There are no meaningfully comparable methods.  However,
sampling error, etc.,  means it is important to have baselines for  ``low'' and ``high'' values.  We use two
baselines.  {\bf Spl:} we evaluate sampling variance in $\fidinf$ by computing the $\fidinf$ between random splits ${\cal O}_i$ of the image set, and obtain
$F_{\mbox{spl}}=\fidinf({\cal O}_i, {\cal O}_j)$.
This yields a ``low'' FID estimate.  {\bf 1D}: to exclude the possibility that our relighting method simply maps pixel values, we train a relighting method using 
only 1x1 convolutions. 

\section{Results}

{\bf Qualitative analysis:}  Figure~\ref{examples} shows five random relightings of five scenes, together with the
relevant multipliers.    These random relights, {\em not curated}, show: shadows removed or replaced;
significant brightening or darkening of surfaces, registered to geometry;
suppression or enhancement of gloss effects; and suppression or enhancement of luminaires.  The model is clearly capable
of substantial, non-obvious relightings that look realistic, and can clearly cope with a very diverse range of surfaces.
Relightings include overall changes of brightness, but are not confined to them.   The model appears to have some form
of geometric knowledge (it often changes the lighting of horizontal and vertical surfaces differently). 

{\bf The uses of LFID:}  If relighting is to be used for data augmentation, one seeks relightings that make large
changes to a scene.  Figure~\ref{bigfig} shows the results of a rejection sampling method that seeks the best LFID for
MSD over a threshold; and the largest MSD for LFID below a threshold.

{\bf Relighting is a successful data augmentation:}  Chogovadze {\em et al} show relighting augmentations improve two
patch matching tasks~\cite{chogovadze2021controllable}.  Table~\ref{classrelight} shows that our relighting augmentations improve
performance on a classification task, with greater impact when the dataset is small (to be expected -- relighted images
contain information from nearest neighbors, redundant when there are enough images).  Improvements for the ``*-room'' categories
exceed those for others, suggesting the augmentation -- which is learned on bedrooms -- generalizes, but not to
arbitrary scenes.  Table~\ref{sslrelight} shows that our relighting augmentations improve performance on a self-supervised learning task.
Note that, while our relighter sees only bedroom images, table~\ref{classrelight} suggests strongly that it produces
acceptable relightings for other scenes, likely a consequence of the common geometry of rooms (see also supplementary).

\begin{table}
  \centerline{
    \footnotesize
  \begin{tabular}{|c||c|c|c|c|}
    \hline
   Egs&Relight&Top-1 acc&TP ``*room''& TP not ``*room''\\
    \hline
    1000&No&51.3&51.2& 51.3\\
    \hline
    1000&Yes&{\bf 51.5}(+0.2)&53.9 (+2.7)&51.4(+0.1)\\
    \hline
    100&No&35.7&37.2&35.6\\
    \hline
    100&Yes&{\bf 36.2}(+0.5)&38.9 (+1.7)&36.0 (+0.4)\\
    \hline
    10&No&24.9&25.8&24.8\\
    \hline
    10&Yes&{\bf 25.5}(+0.6)&27.1(+1.3)&25.4 (+0.6)\\
    \hline
  \end{tabular}}
  \caption{\em  Relighting augmentations are helpful in a classification task.  We use an imagenet pretrained resnet 50
    feature stack (cf~\cite{he2016deep}) to classify images from places365 (\protect \cite{zhou2017places}; 365 classes).  For each case, we
    use the given number of randomly selected examples {\em per class} (at 10 per class, overfitting is pronounced; we
    average over multiple training runs to suppress variance); augment with  crop, flip, color jitter and, if using relighting,    relight[alpha=0.1].  Note the effect of relighting is more pronounced with few examples.  There are 28 classes named
    ``*room''; notably, the test precision (TP) for these classes improves more than for others.  
    \label{classrelight}
  }
  \end{table}
 
\begin{table}
  \centerline{
    \footnotesize
  \begin{tabular}{|c||c|c|}
    \hline
Augs&Top-1 acc & Top-1 acc relight\\
   \hline
  A&16.1&{\bf 16.3}\\
 B&16.1&{\bf 16.5}\\
 C&14.2&{\bf 15.2}\\
 \hline
  \end{tabular}}
  \caption{\em  Relighting augmentations are helpful in a self-supervised learning task.  We train a feature stack
    using SIMCLR \protect \cite{chen2020simple} with three different  augmentation conditions: A - original from the paper; B -
    without left/right flip; C crop with scale in [0.36, 1].  For relight, we use pairs of images with distinct relights. We
    train on 10K LSUN bedroom images, freeze, then train a linear readout on places365 36K; then evaluate.  Note that
    using relighted images reliably offers a small gain.
    \label{sslrelight}
  }
  \end{table}

\begin{figure}
  \centerline{\includegraphics[width=\columnwidth]{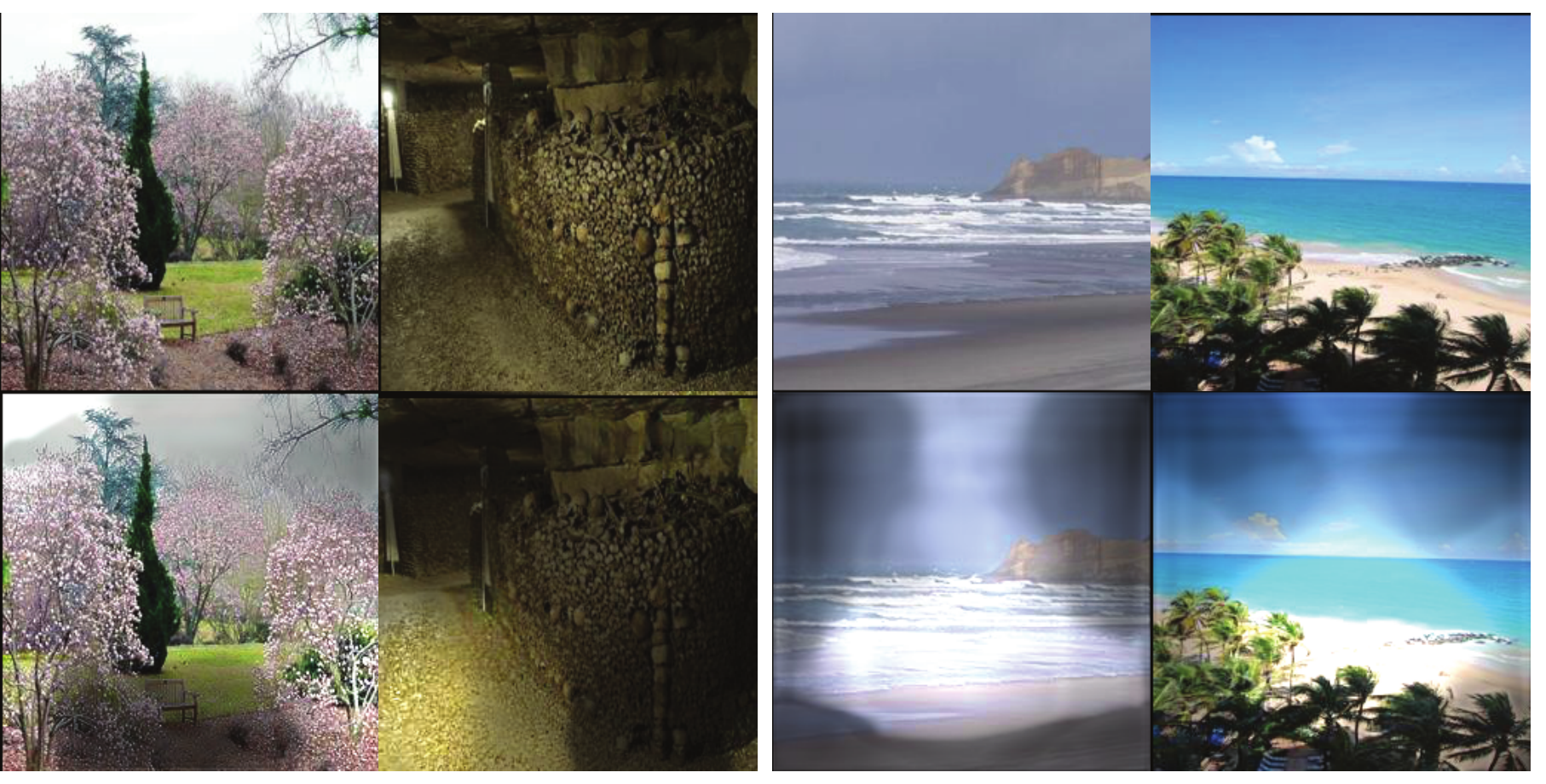}}
  \caption{\em Rough shape -- rather than scene label -- seems to be key to this method's success. {\bf Left:} images ({\bf top}) from
    Places365 categories {\em botanical\_garden} and {\em catacomb} and relightings ({\bf bottom}); because the
    rough shape is like that of a bedroom, the relightings are quite good.  {\bf Right:} images from {\em ocean}; relightings are
    inaccurate, likely because the model cannot deal with this class of spatial layout; note how the relightings echo the geometry of a room.
  \label{general}  }
\end{figure}

\section{Discussion}\label{discussion}


Further work will likely include: using these results to regularize PRT estimation; investigating the effects of
architectural choice on the network; and reestimating nearest neighbors with generator estimates.
{\bf Limitations:} Some relightings are clearly not present, and generalization to other scenes is variable (Figure~\ref{general}).  We have no procedure for improving the model, save
collecting more (unlabelled) data.

{\bf Negative social implications:}  Convincing fake images can pollute (say) the news media; but our method offers only
relighting, rather than the much more dangerous insertion or deletion editing.
Experiments required approx 250 hours of time on a single GTX3090 from method inception to submission.  The datasets used either do not contain, or are not known to contain, personally identifiable or offensive material.  A total of \$150 was paid to Amazon mechanical Turk workers for image labelling; no
personally identifiable information was collected.

\section*{Appendix I:  Further variants and fuller comparisons}
\begin{figure*}[ht!]
\centerline{\includegraphics[width=\textwidth]{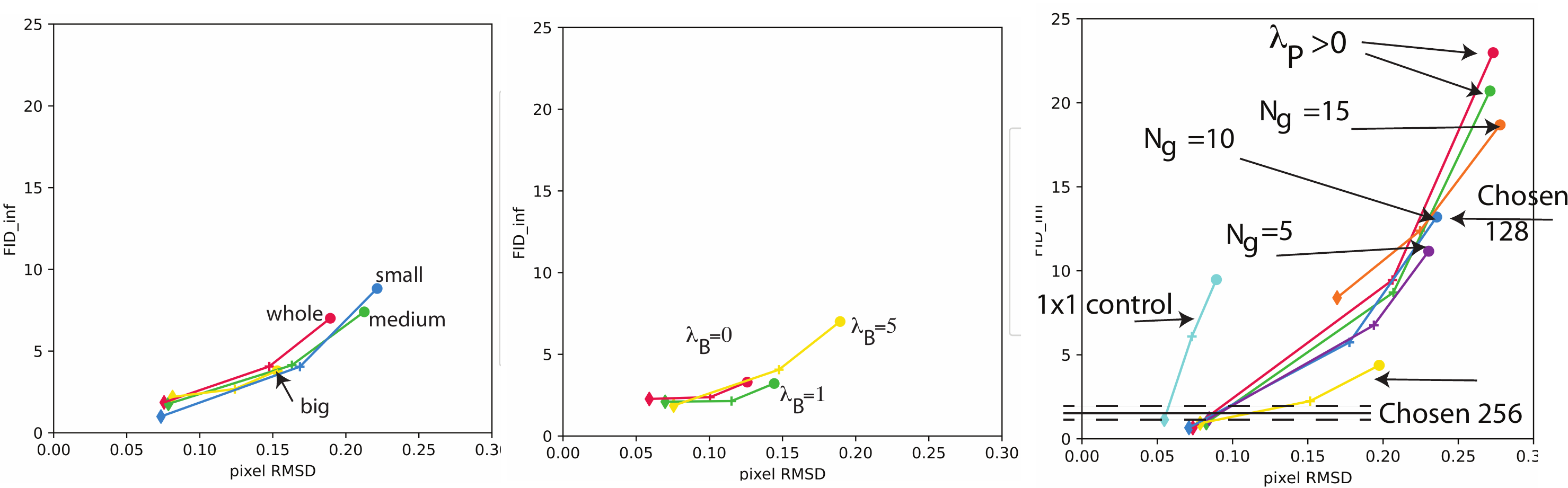}}
\caption{\em
1  $\fidinf$ against MSD for various parameter choices for model selection.  {\bf Left:} PatchGAN adversary scale affects
  performance; the {\bf small} receptive field is best.  {\bf Center:} $\lambda_B$ encourages variations without much adverse affect on $\fidinf$;
  we chose $\lambda_B=5$.  {\bf Right:}     $\lambda_P>0$ reliably produces more speculative models with weaker FID; the figure is
  Figure 4 from paper, for reference.
      \label{bpscale}
  }
\end{figure*}

\begin{figure}[h!]
\centerline{\includegraphics[width=\columnwidth]{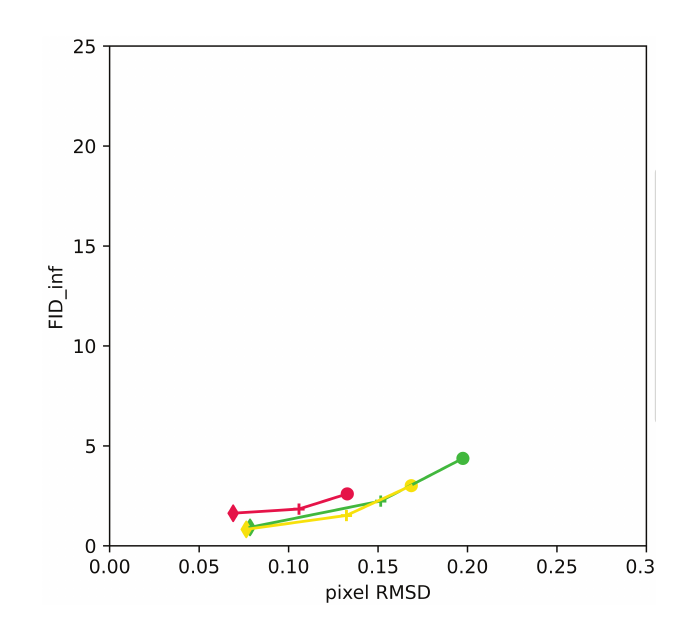}}
  \caption{\em  Averaging for equivariance during inference hurts metrics.  $\fidinf$ against MSD for {\bf red:} 256x256 images relighted using averaging at inference time; {\bf green:} relighted 256x256 square crops; {\bf yellow:} 256x256 square crops from averaged images.  
    \label{averaging}
  }
\end{figure}

\begin{figure*}[h!]
\centerline{\includegraphics[width=\textwidth]{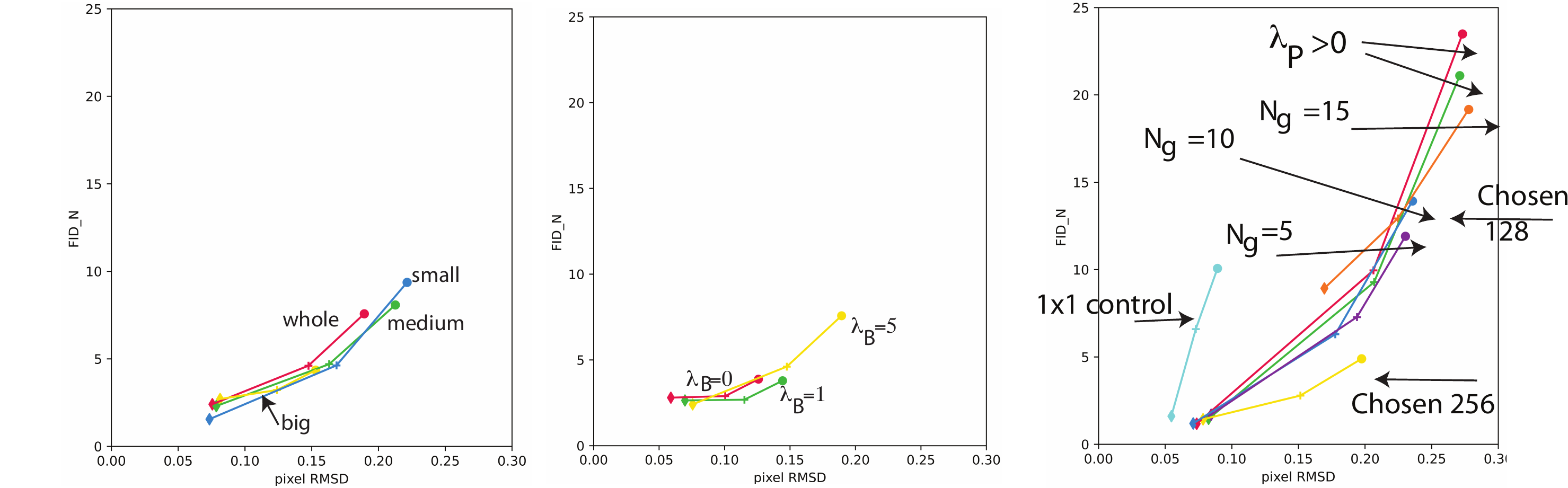}}
\caption{\em
No major effects result from using FID rather than $\fidinf$.  FID  against MSD for various parameter choices for model selection.  {\bf Left:} PatchGAN adversary scale affects
  performance; the {\bf small} receptive field is best.  {\bf Center:} $\lambda_B$ encourages variations without much adverse affect on $\fidinf$;
  we chose $\lambda_B=5$.  {\bf Right:}     $\lambda_P>0$ reliably produces more speculative models with weaker FID.  Compare Fig.~\ref{bpscale}.
      \label{bpscale-fid}
  }
\end{figure*}


{\bf Equivariance:}  A function $\phi: {\bf x} \in X \rightarrow {\bf y} \in Y$ is equivariant under the action of a group $G$ if there
are actions of $G$ on $X$ and $Y$ such that $\phi(g \circ {\bf x})=g \circ \phi({\bf x})$. Information being
gained or lost at the boundary is an obstacle to applying the theory of group actions exactly (except for certain finite
groups~\cite{CohenWelling}).   If one relaxes the definition to require only an approximate match, well-known visual
feature representations tend to have strong equivariance properties either by design or in practice~\cite{LencVedaldi}.
Imposing equivariance properties on intrinsic image decompositions (albedo at a point should not depend on image crop)
shows sharp improvements \cite{DAFTPAMI}.  Relighting has a form of equivariance
property: small movements of the camera frame should not affect the overall relighting of a scene (apart from rows and
columns of the image moving in and out of view).  We use that paper's crude averaging strategy because we know no better.
In this approach we: pad an image; estimate multipliers for a number of closely overlapping crops of the padded image; average overlapping estimates to obtain final multipliers.  Doing so reliably loses both MSD and $\fidinf$, for reasons we do not know (Figure~\ref{averaging}).

  {\bf Pixel uniformity loss:} We expect there is some lighting that makes any location look bright (meaning the image value at each
  location should be bright), and that any location might be in darkness (meaning that the shading value at each location might be
  dark).  Write $\uparrow{M_V I_V}$ for the largest value at each location over all generators of $M_V I_V$ and
  $\downarrow{M_V S_V}$ for the smallest value at each location over all generators of $M_V S_V$.  We use:
  \[
    {\cal L}_{P}(I_V; \phi)=\overline{E}
    \]
    where
    \[E= \left[-\mbox{min}(0.95-\uparrow{M_V I_V}, 0)+
      \mbox{max}(\downarrow{M_V S_V}-0.05, 0)\right]
    \]
    Pixel uniformity loss reliably produces more speculative models with weaker $\fidinf$ (Figure~\ref{bpscale}).


\begin{figure*}[h!]
  \centerline{\includegraphics[width=\textwidth]{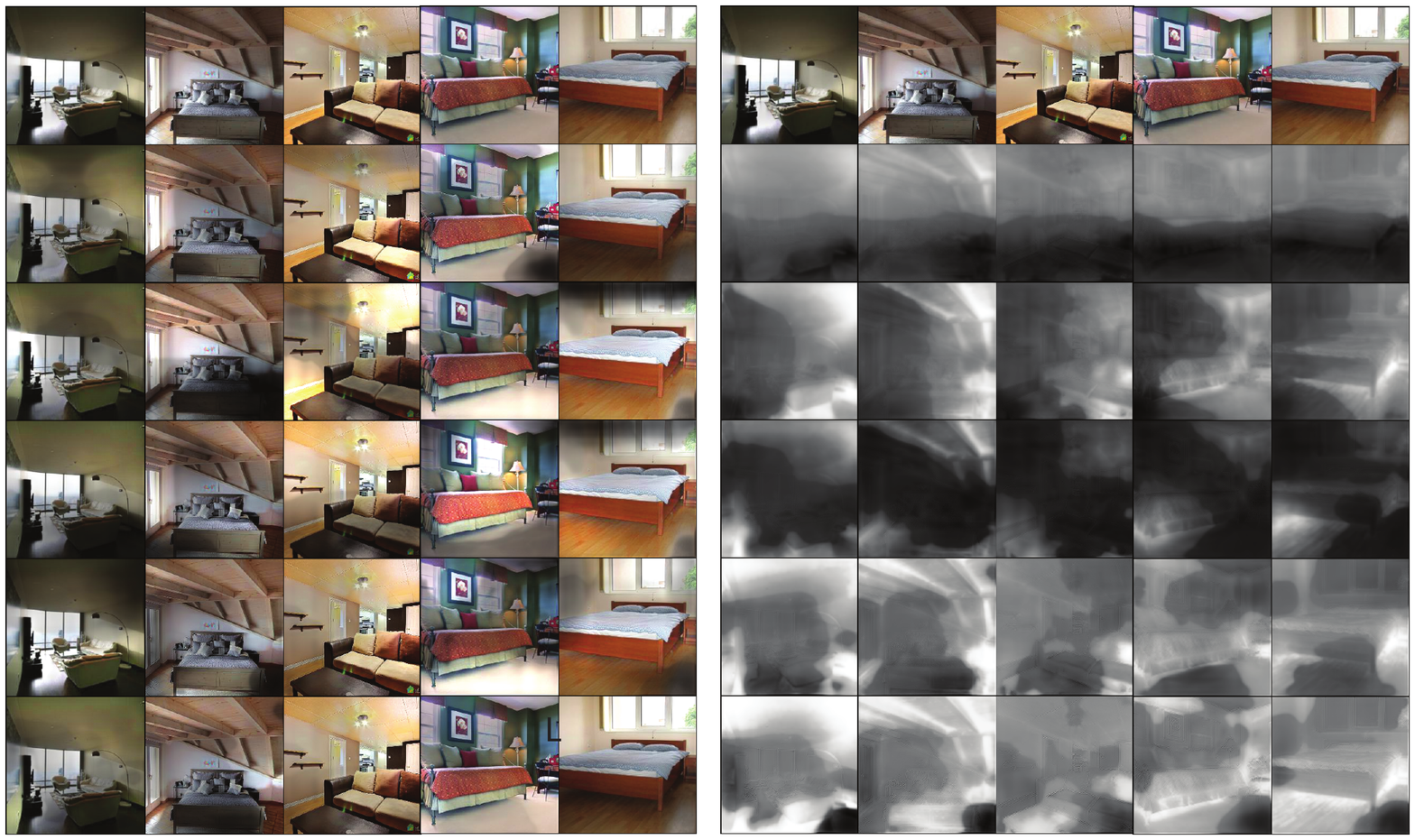}}
\caption{\em
  Images relighted at $\alpha=0.1$ (more variable; {\bf left}; cf  $\alpha=0.5$, less variable, $\alpha=0.01$, much more variable, in Fig~\ref{p5p01}) and the first five multiplier fields for
  each {\bf right}.  Notice how the multipliers are keyed to structures in the image -- the method ``knows'' some
  geometry and radiometry. Reproduced from Fig 3 of main paper, without arrows, for reference.
  \label{examples}
  }
\end{figure*}

\begin{figure*}[h!]
  \centerline{\includegraphics[width=\textwidth]{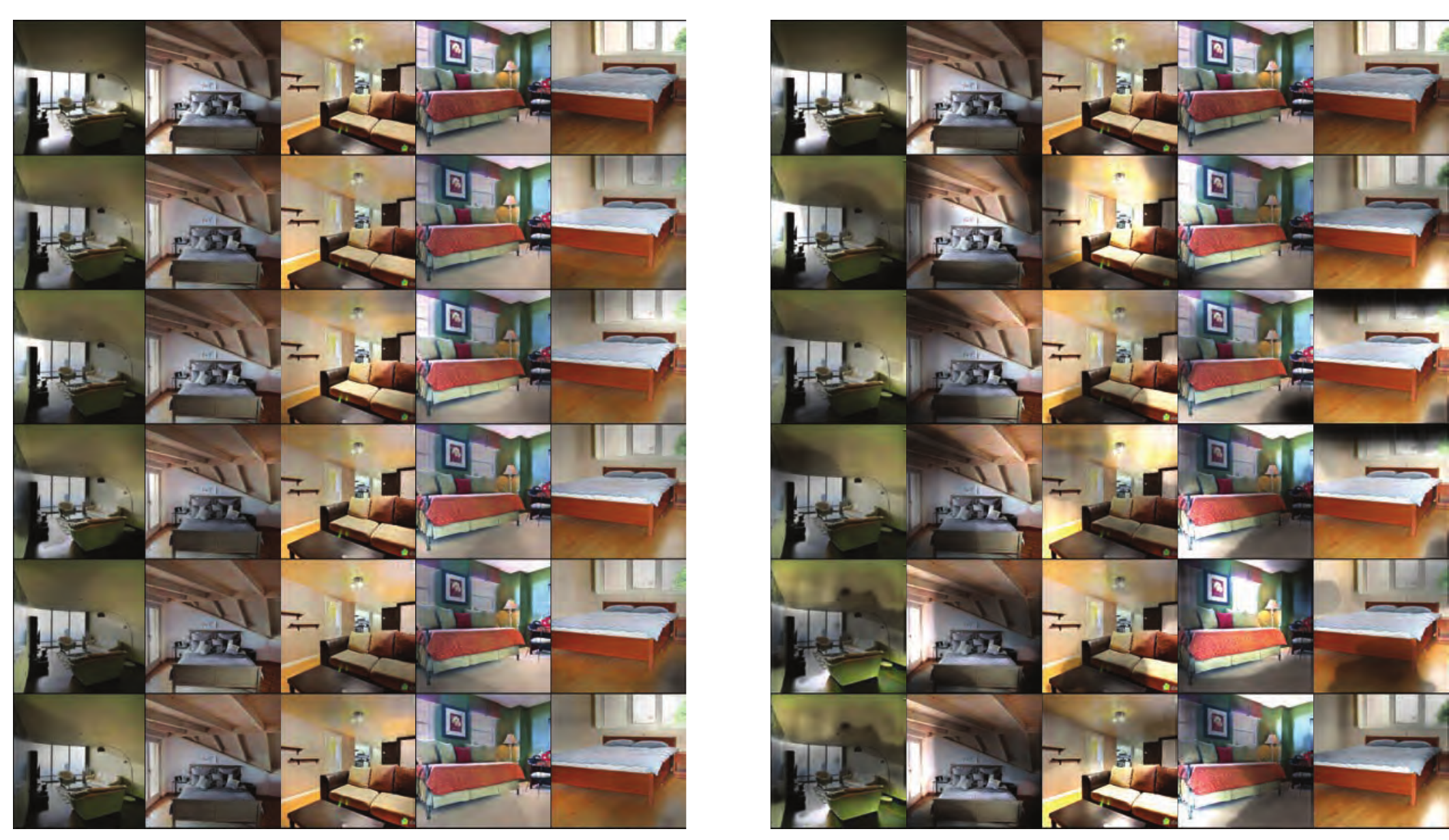}}
\caption{\em
  Images relighted at $\alpha=0.5$  {\bf left},  less variable and  $\alpha=0.01$, much more variable {\bf right}.
  At $\alpha=0.5$ images have higher quality, but tend to be less variable.  At $\alpha=0.01$, the difficulties presented
  by extrapolation become more obvious.
  \label{p5p01}
  }
\end{figure*}

\begin{figure*}[h!]
\centerline{\includegraphics[width=\textwidth]{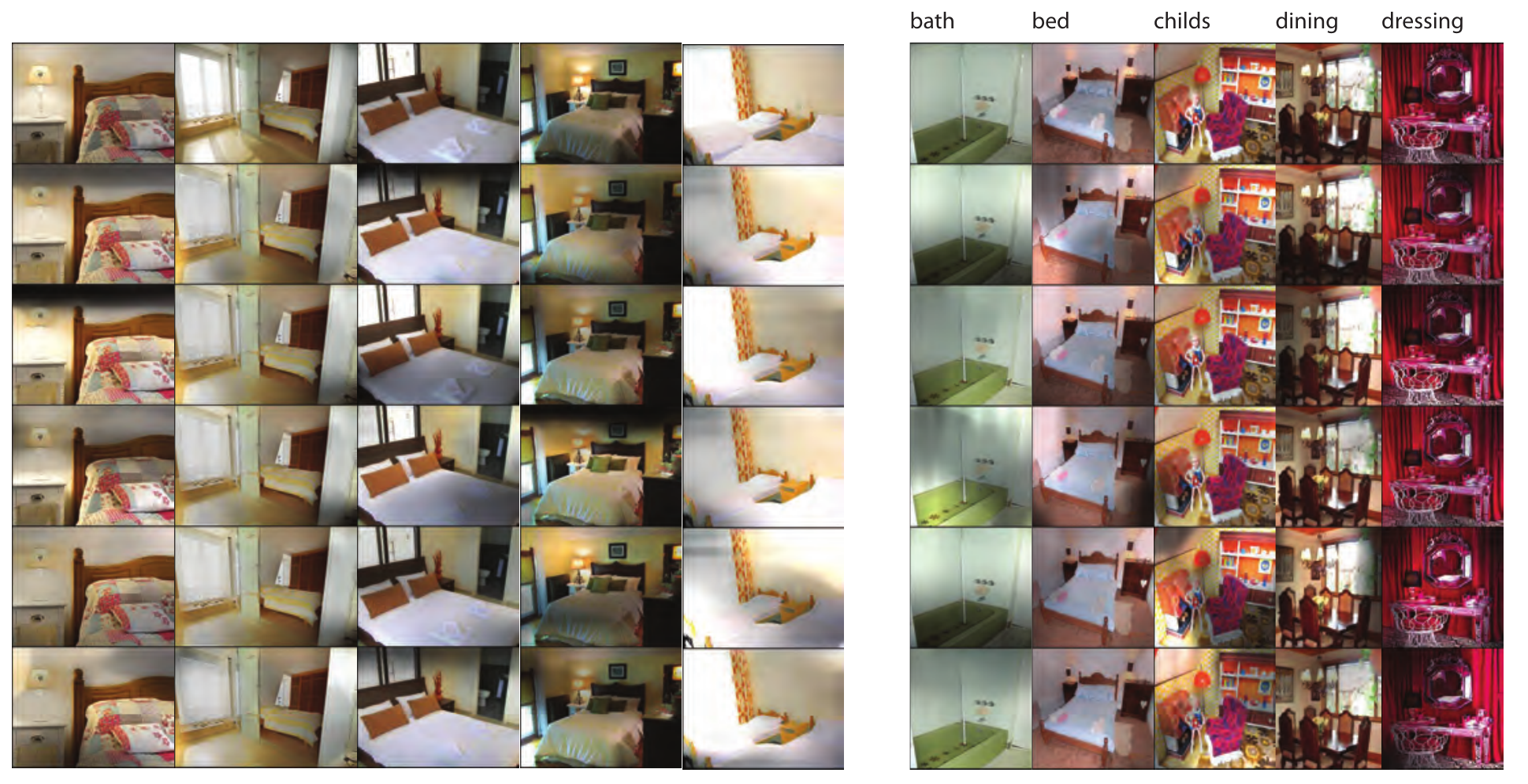}}
\caption{\em Our method produces qualitative changes in the distribution of light, and  appears to generalize well across datasets. {\bf Left:} Further qualitative results on LSUN-Bedrooms; {\bf Right:} Results for the first image each of the validation
  set for 10 scenes with ``room'' in their name from the PlacesRooms dataset; relights are {\em not} curated.
  {\bf Top:} Image; {\bf other:} relights obtained using our method, $\alpha=0.5$.
      All relights are shown at the same mean intensity as the original image, to ensure that effects are not masked by
      overall shift in brightness.
      \label{gen}
  }
\end{figure*}

\begin{figure*}
\centerline{\includegraphics[width=\textwidth]{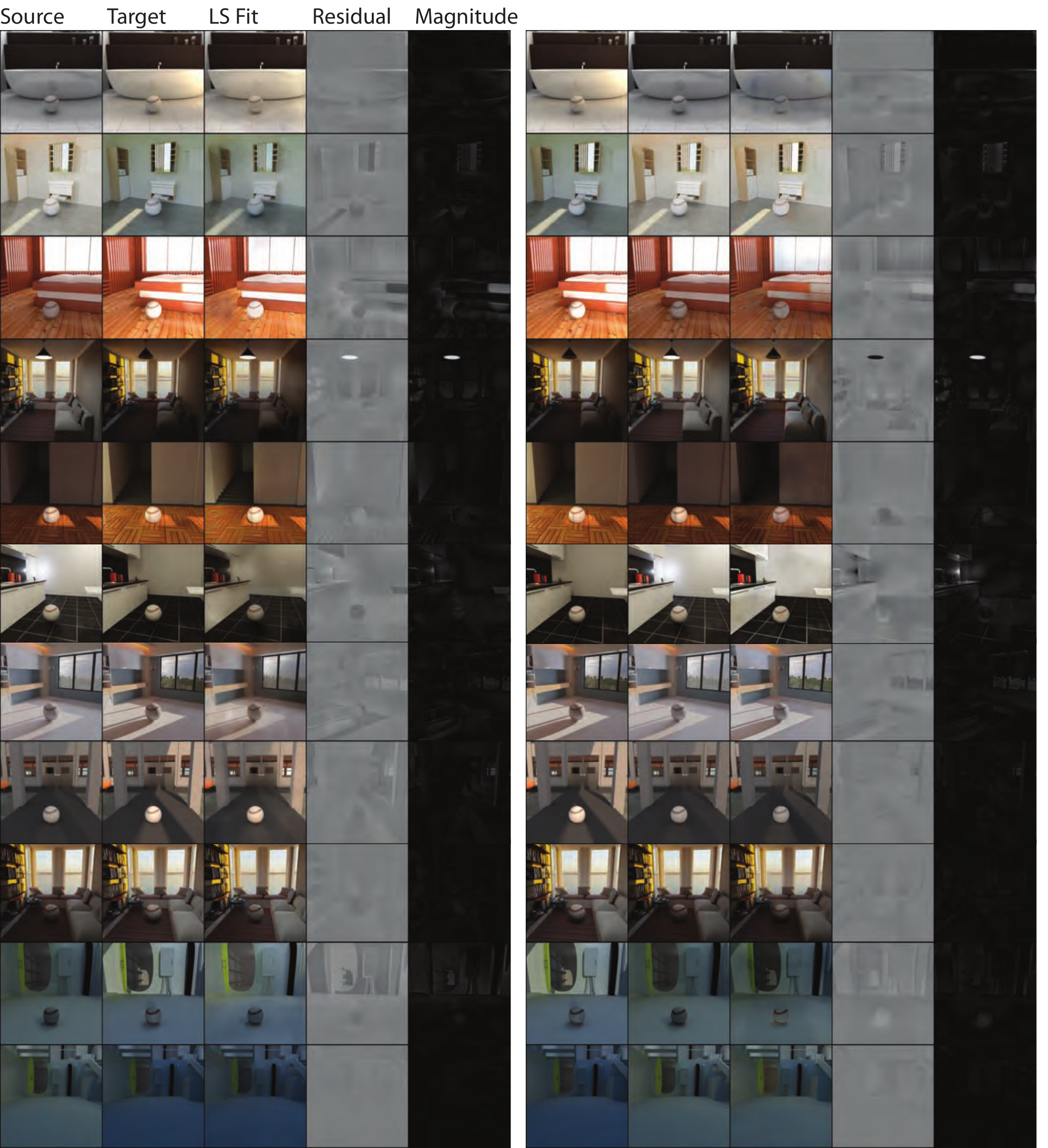}}
\caption{\em
  We show all pairs of source/target/least-squares images used to prepare Figure 5 of the main paper.  Recall that shows
results of: rendering CGI scenes under two lights; computing the
  multipliers for the {\bf source}; constructing weighted sum of multipliers that yields a least squares fit to the {\bf
    target}; then comparing target to the {\bf least squares}.  The {\bf residual} error in intensity
  is shown on a scale where white is 1; mid grey is 0; and black is -1.  Note relatively good fits, suggesting that
  for the cases we have investigated, multipliers are a reasonable fit.  {\bf Quantitative:} over a total of 11 pairs
  (22 tests), for a dataset we will provide, we find average metrics between the least squares prediction and the target of
  RMSD:   0.0503  PSNR: 26.6  LPIPS: 0.0648, suggesting quite successful prediction of the target
  by source multipliers.
      \label{bryan}
  }
\end{figure*}

\clearpage
\section{Appendix II: Proof of Theorem 1}

We wish to compare the radiosity of a scene with that of a similar scene, of somewhat different geometry, albedo and luminaire.
The key result is the difference between these two radiosities can be described in a sensible way, and can
be bounded.  In turn, it follows that sufficiently similar scenes will share generators.

\subsection{Notation}

It is enough to work with a simplified model of a scene $V$ as a tuple $({\cal G}, \rho, E)$ of geometry, albedo and luminaire (rather than stochastic family of luminaires).  ${\cal G}$ consists of a set
of surfaces parametrized in some way and a 2D domain for that parametrization; write ${\cal G}=(\vect{s}(\vect{x}), {\cal
  D})$.  Significant complications in the analysis can occur associated with concave ``creases'' or at surface interpenetrations. 
These complications aren't physically important -- every corner in a room is made smooth by at least a thin layer of paint or
dirt -- and so we avoid them by assuming:
\begin{itemize}
\item ${\cal D}$ is compact (but need not be connected) with finite area.
\item Each connected component of ${\cal D}$ represents a distinct surface.
\item Surfaces are parametrized by $\vect{s}(\vect{x})$, where $\vect{s}$ is at least $C^1$ and is an embedding
  (surfaces do not self-intersect).  
\item Surfaces are orientable, and are oriented.
\item For any pair of points $\vect{x}$, $\vect{y}$, we have $k(\vect{x}, \vect{y}) \leq C$ for some large finite constant $C$ and $k$ defined  below.  This rules out ``throats'' -- places where the gap between surfaces narrows and then opens out, Figure~\ref{notation} -- in the geometry, and bounds the largest curvatures.   Larger $C$ corresponds to narrower throats and tighter curvatures being allowed.  The condition is sufficient to ensure the kernel is compact, below. 
\end{itemize}
  Note that parametrizing surfaces by atlases of charts complicates notation
  without changing the analysis; we will suppress these details.   Write
\[
  \pnorm{E}=\left[\int_{\cal D} E^p d\vect{x}\right]^{1/p}
\]
($1\leq p \leq \infty$) for the $p$ norm of $E$.  Recall that for an operator ${\bf K}$
\[
  \pnorm{{\bf K}}\leq \linfnorm{{\bf K}}^{\frac{(p-1)}{p}} \gennorm{{\bf K}}_1^{\frac{1}{p}}
  \]
  (eg~\cite{Arvo},~\cite{Kato} p.144 or~\cite{Dunford} p.518).  All operators here will have both $\gennorm{{\bf K}}_1<1$ and $\linfnorm{{\bf K}}<1$,
so that results apply for any $p$; we write $\gennorm{{\bf K}}$.

We wish to compare the radiosity $B_V$ of a scene $V$ with the radiosity $B_V'$ of a similar scene $V'$.  These scenes have somewhat different geometry, albedo and luminaire.
We consider affine transformations of the geometry, so that ${\cal G}'({\cal G}, \matx{A}, \vect{b})=(\matx{A} \vect{s}(\vect{x})+\vect{b}, {\cal D})$.
We will prove:

{\bf Theorem 1:} {\em For $V=({\cal G}, \rho, E)$ and $V'=({\cal G}'({\cal G}, \matx{A}, \vect{b}), \rho', E')$, where
  $\epsilon_E\gennorm{E} = \gennorm{E-E'}$, $\epsilon_\rho=\mbox{sup}_{\cal D}\abs{\left[\rho-\rho'\right]}$, $p=\mbox{sup}_{\cal D} \rho$,
  $p'=\mbox{sup}_{\cal D}\rho'$, $c$ is the condition number of $\matx{A}$ (ratio of largest to smallest eigenvalues), 
  we have:}
\[
\gennorm{B_{V}-B_{V'}}\leq\left[\begin{array}{c}
    \frac{\epsilon_E}{1-p}+ \\
    \left[\frac{\epsilon_\rho(1+\epsilon_E)}{(1-p)(1-p')}\right]+\\
    \frac{(c^4-1)(1+\epsilon_E)}{(1-p')^2}\end{array}\right]  \gennorm{E}
\]

\subsection{Standard results}

\begin{figure*}[ht!]
  \centerline{
    \includegraphics[width=0.5  \textwidth]{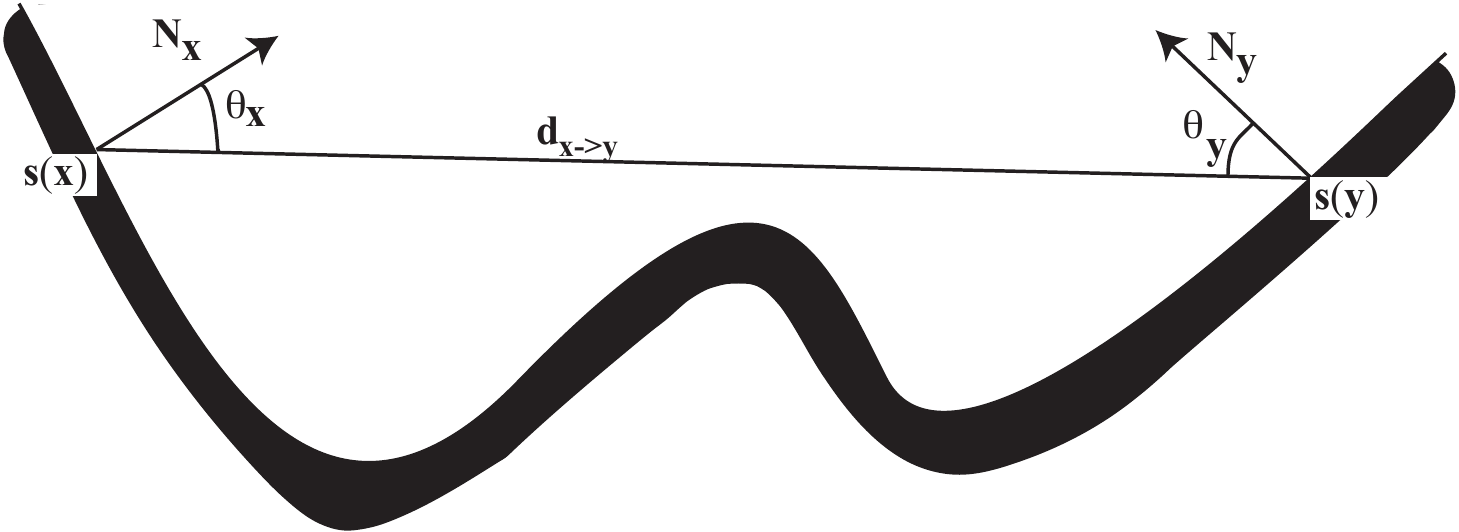}
   \includegraphics[width=0.5  \textwidth]{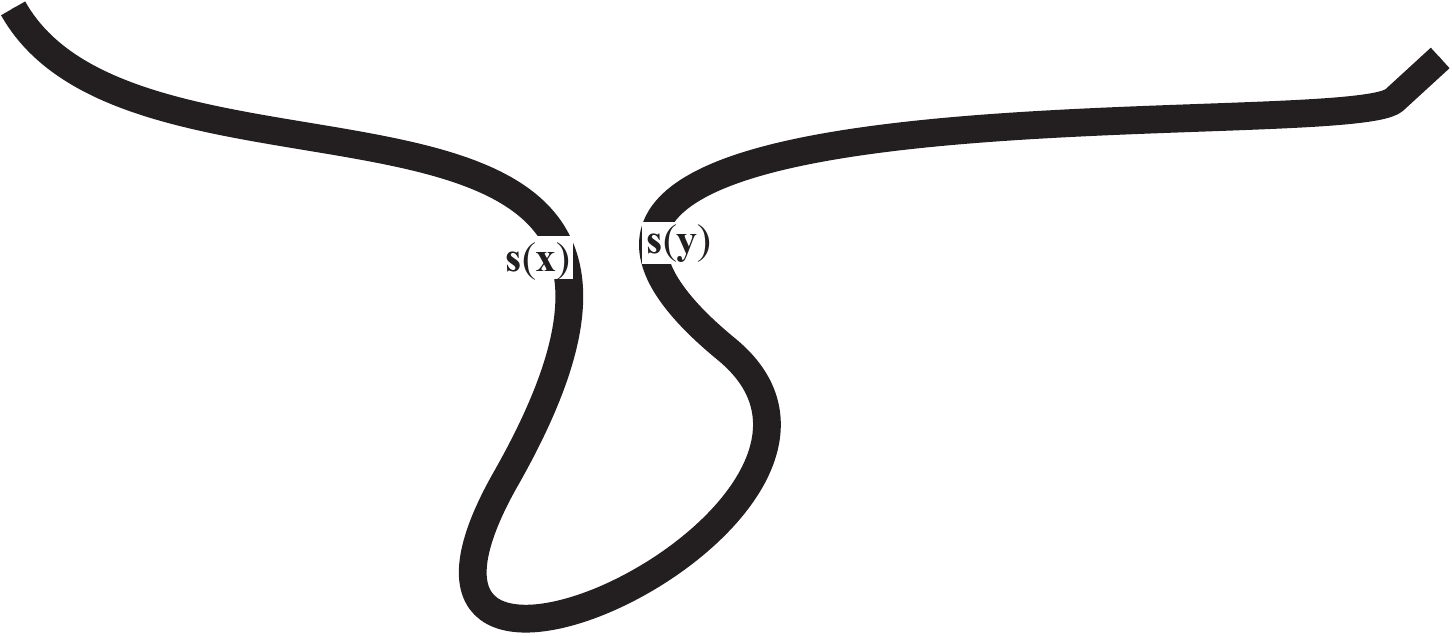}    
    }
\caption{\em {\bf Left:} Our notation.  {\bf Right:} a throat.
\label{notation}}
\end{figure*}

 The geometry consists of a set of parametrized surfaces.  Write ${\cal D}$ for the domain of the parametrization.
For a parameter vector $\vect{x}\in {\cal D}$ write $\vect{s}(\vect{x})$ for the corresponding surface point; $B(\vect{x})$ for radiosity at that point; $\rho(\vect{x})$ for the albedo at that point; 
$E(\vect{x})$ for the radiant exitance at that point; $V(\vect{x}, \vect{y})$ for the {\em visibility function} that is $1$ when there is a direct line of sight from $\vect{s}(\vect{x})$ to
$\vect{s}(\vect{x})$ and 0 otherwise; $\vect{d}_{\vect{x}\rightarrow \vect{y}}=\vect{s}(\vect{y})-\vect{s}(\vect{x})$; and
$\vect{N}(\vect{x})$ for a unit normal at $\vect{x}$.
Figure~\ref{notation} shows some notation commonly used.
Standard diffuse interreflection theory (see,
eg~\cite{nn36019,DAFAZ2,SillionBook,CohenBook})
gives
\[
  B(\vect{x})=E(\vect{x})+\rho(\vect{x}) \int_{\cal D}
  k(\vect{x}, \vect{y}) B(\vect{y}) dA_{\vect{y}}
\]
where
\begin{eqnarray*}
  k(\vect{x}, \vect{y})&=&\frac{1}{\pi} \frac{  V(\vect{x}, \vect{y})\cos \theta_{\vect{x}}\cos \theta_{\vect{y}}}{\ltwonormsq{\vect{d}_{\vect{x}\rightarrow\vect{y}}}}\\
  &&\mbox{\hspace{0.5cm} (notation in Figure~\ref{notation})}\\
  &=&
  \frac{  V(\vect{x}, \vect{y})
    \left[\vect{N}^T(\vect{x})\vect{d}_{\vect{x}\rightarrow \vect{y}}\right]
    \left[\vect{N}^T(\vect{y}) \vect{d}_{\vect{y}\rightarrow \vect{x}}\right]
    }
    {
      \left[\vect{d}^T_{\vect{x}\rightarrow \vect{y}}\vect{d}_{\vect{x}\rightarrow \vect{y}}\right]^2
    }
\end{eqnarray*}
(recalling $\vect{N}$ is a unit normal).  Write  ${\bf D}_\rho$ for the linear operator that maps $f(\vect{x})$ to
$(\rho f)(\vect{x})$.   We can now abbreviate into $B=E+ {\bf D}_\rho {\bf T}  B=E+{\bf K} B$,
where ${\bf D}_\rho$, ${\bf T }$ and ${\bf K}$ are linear integral operators.
It is known that
\[
\gennorm{{\bf T}}\leq 1
\]
(for example, our ${\bf T}$  is the ${\bf K} {\bf G}$ of~\cite{Arvo}, sec 5.3, with constant diffuse BRDF of $1$).
Furthermore, writing
\[
p=\linfnorm{\rho(\vect{x})}<1
\]
we have 
\[
\gennorm{{\bf D}_\rho}\leq p
\]
by simple calculation and so
\[
\gennorm{{\bf K}}\leq p.
\]
For our assumptions,
\[
\int_{\cal D} \int_{\cal D} \mid \! k(\vect{x}, \vect{y})\! \mid^2 dA_{\vect{x}} dA_{\vect{y}}\leq C^2 \mbox{Area(${\cal D})$}^2,\]
so that $k$ is a Hilbert-Schmidt kernel.
Because the kernel is Hilbert-Schmidt, ${\bf K}$ is a compact operator (eg~\cite{Conway}).
There is then  a Neumann series solution
\begin{eqnarray*}
  B&=&{\bf M}^{-1}E\\
  &=&E+{\bf K} E+ {\bf K}^2 E+ \ldots\\
  &=&\mbox{exitance}+\mbox{one bounce}+\mbox{two bounces}+\ldots
  \end{eqnarray*}
  which converges for all $r<1$ (and so all practical scenes).  It follows that
  \[
    \gennorm{{\bf M}^{-1}} \leq \frac{1}{1-p}
    \]

\subsection{Lemma 1: Perturbed Luminaire}

Assume that the luminaires for $V'$ differ from those of $V$ (but geometry and albedo are the same), so $V=({\cal G},
\rho, E)$ and $V'=({\cal G},\rho, E')$. Standard results yield:

\vspace{5mm}
\noindent  {\bf Lemma 1:} {\em  using the notation above}
\[
  \gennorm{B_V-B_{V'}}\leq \frac{1}{1-p} \gennorm{E-E'}=\frac{\epsilon_E}{1-p} \gennorm{E}.
  \]

  \vspace{5mm}
  \noindent {\bf Proof:} $\gennorm{{\bf M}^{-1}}$ is bounded by $\frac{1}{1-p}$, above. $\square$ 
 \vspace{5mm}
  
\subsection{Lemma 3:  Perturbed Albedo}

Now assume that $V=({\cal G},\rho, E)$ and $V'=({\cal G},\rho', E)$ and 
$\rho'=\rho+\delta$.  Write $\epsilon_\rho=\gennorm{{\bf D}_\delta}$.  We expect $B'$ to differ
from $B_{V}$ because it contains multiple bounce terms (above).  For example, the shading in a very dark room is notably
different from that in a very light room (eg~\cite{nn36019,DAFAZ2,Gilchristbook}), because low albedos mean that the multiple bounce
terms are very small.  Equivalently ${\bf M}$ and ${\bf M}'$ are
different.   However, ${\bf G}$ is bounded and we have
$\gennorm{{\bf M}}=p\leq1$ and $\gennorm{{\bf M}'}=p' \leq1$.  The dynamic range of practical albedos is
quite limited (eg~\cite{Brelstaff}), so it is reasonable to bound $\rho(\vect{x})$ away from both 0 and 1.

\vspace{5mm}
\noindent {\bf Lemma 2:} {\em using the notation above, for all pairs of $0<\epsilon \leq \rho \leq (1-\epsilon)<1$ and
  $0<\epsilon \leq \rho'\leq (1-\epsilon) <1$, $\gennorm{{\bf D}_{\rho-\rho'}}=\epsilon_\rho < 1$}

\vspace{5mm}
\noindent {\bf Proof:} for some test function $f$,
\[{\bf D}_{\rho-\rho'} f=(\rho(\vect{x})-\rho'(\vect{x})) f(\vect{x}).
\]
Then
\begin{eqnarray*}
\lonenorm{{\bf D}_{\rho-\rho'} f}&=&\int_{\cal D} \mid\!(\rho(\vect{x})-\rho'(\vect{x})) f(\vect{x})\!\mid dA_{\vect{x}}\\
  &\leq&\int_{\cal D}\mid\!(\rho(\vect{x})-\rho'(\vect{x}))\!\mid \mid\! f(\vect{x})\!\mid dA_{\vect{x}}\\
&\leq&\dafsup{\vect{x}}\left[\mid\!(\rho(\vect{x})-\rho'(\vect{x}))\!\mid\right] \int \mid\! f(\vect{x})\!\mid dA_{\vect{x}}.\\
\end{eqnarray*}
The constraints on $\rho$, $\rho'$ ensure
\[
\sup{\vect{x}}\left[\mid\!(\rho(\vect{x})-\rho'(\vect{x}))\!\mid\right]<1
\]
and $\linfnorm{{\bf D}_{\rho-\rho'}}<1$, so the result follows for all norms.
$\square$
\vspace{5mm}

This yields the novel (but fairly straightforward)

\vspace{5mm}
\noindent {\bf Lemma 3:} {\em using the notation above}
\[  \gennorm{B_{V'}-B_V} \leq \left[\frac{1}{1-p}\right]\left[\frac{1}{1-p'}\right] \epsilon_\rho \gennorm{E} \]

\vspace{5mm}
\noindent {\bf Proof:} We have $(I-{\bf D}_\rho {\bf G})  B_{V}=E=
(I-{\bf D}_{(\rho+\delta)} {\bf G})  B_{V'}$, so
\begin{eqnarray*}
  B_{V'}-B_{V}&=&{\bf D}_{\rho'}{\bf G} B_{V'}- {\bf D}_{\rho}{\bf G} B_{V}\\
      &=&{\bf D}_{\rho'}{\bf G} B_{V}- {\bf D}_{\rho}{\bf G} B_{V}+{\bf D}_{\rho'}{\bf G} (B_{V'}-B_{V})\\
              &=&{\bf D}_{\delta}{\bf G} B_{V}+{\bf D}_{\rho'}{\bf G} (B_{V'}-B_{V})\\
              &=&{\bf D}_{\delta}{\bf G} {\bf M}^{-1}E+ {\bf D}_{\rho'}{\bf G} (B_{V'}-B_{V})
\end{eqnarray*}
so
\[
  B_{V'}-B_{V}={\bf M}'^{-1}{\bf D}_{\delta}{\bf G} {\bf M}^{-1}E.
\]
Now $\gennorm{{\bf D}_\delta {\bf G}}\leq\epsilon_\rho$, so
\begin{eqnarray*}
  \gennorm{B_{V'}-B_{V}}&=&\gennorm{{\bf M}'^{-1}{\bf D}_\delta {\bf G} {\bf M}^{-1}E}\\
                            &\leq& \left[\frac{1}{1-p}\right]\left[\frac{1}{1-p'}\right] \epsilon_\rho \gennorm{E}
\end{eqnarray*}
$\square$
\vspace{5mm}

\subsection{Theorem 1:  Perturbed Geometry}

We wish to study the effect of small changes in geometry on the solution $B$. 
We have found no results on the effect of perturbing geometry, which appear to be hard to obtain (cf~\cite{Arvothesis}, p176 says they are;~\cite{JacKupp} estimate the effect of flat ports on an integrating sphere).

Small geometric changes can produce major changes in
visibility which are difficult to bound.   We consider affine transformations of the 
surfaces, so that $\vect{s}(\vect{x})$ is mapped to ${\cal T}(\vect{s})=\matx{A} \vect{s}(\vect{x})+\vect{b}$ ($\mbox{det}(\matx{A})\geq0$). 
We assume that the smallest singular value of $\matx{A}$ is positive.
There are two advantages of this model: first, it is a reasonable description of small changes between otherwise quite
similar rooms (small anisotropic scalings in the right coordinate system); second, the visibility function for ${\cal
  T}(\vect{s})$ is the same as the visibility function for $\vect{s}$ (lemma 4, below).  

\subsubsection{Transformations and Functions}

The transformation produces a natural bijection between functions of
original and transformed geometry.  This is so natural that it can be
difficult to see at  first glance.  Assume we have a function $f$ defined on the first geometry.  We must know the value
that $f$ takes at each point on the geometry, which means we know the value at each point in the parameter space.  But
this parameter space parametrizes the transformed geometry, too, so we have a corresponding function on the transformed
geometry (this argument yields that the mapping between function spaces is a bijection).  For $f$ a function defined on
the original parametrization,  write ${\cal T}(f)$ for the transformed version. 

This natural correspondence may not respect the geometric transformation on the embedded surfaces.  This means
we need to distinguish between a function mapped through the correspondence
and one evaluated on the new geometry.  For example, consider the function
$\mbox{sqdist}(\vect{x}_0,\vect{x}_1)=\ltwonormsq{\vect{s}(\vect{x}_0)-\vect{s}(\vect{x}_1)}$.  If $\matx{A}=2*Id$,
the distance between points on the transformed geometry has doubled.  Write $\mbox{sqdist}'(\vect{x}_0, \vect{x}_1)=\ltwonormsq{\matx{A}\vect{s}(\vect{x}_0)-\matx{A}\vect{s}(\vect{x}_1)}$ (the prime indicates the function is evaluated on the transformed geometry).   We have ${\cal
  T}(\mbox{sqdist})(\vect{x}_0,\vect{x}_1)=(1/4) \ltwonormsq{\matx{A}\vect{s}(\vect{x}_0)-\matx{A}\vect{s}(\vect{x}_1)}=(1/4) \mbox{sqdist}'(\vect{x}_0,\vect{x}_1)$.

\subsubsection{Transformations and the Kernel}

A similar natural correspondence applies to operators.  We must now consider the relationship between
${\cal T}({\bf K})$  and ${\bf K'}$.  Each is a bounded linear operator defined on functions of ${\cal G}'$, but they are different.
${\cal T}({\bf K})$ is obtained by applying the natural correspondence to ${\cal K}$;  in contrast ${\cal K}'$ is obtained by computing the
kernel {\em using the geometry of ${\cal G}'$}, which may give a different operation.

By tracking the effects of affine transformations on the terms of $k$, we can constrain the difference.  First, we have\\

\vspace{5mm}
\noindent {\bf Lemma 4:} {\em  $V(\vect{x}, \vect{y})={\cal T} (V(\vect{x}, \vect{y}))=V'(\vect{x}, \vect{y})$}\\

\vspace{5mm}
\noindent {\bf Proof:} The main issue is keeping track of notation; the first equality is obvious (natural correspondence);
the second -- that computing the view function in the transformed space is equivalent to computing it in the original
space, and mapping through correspondence -- follows because affine transformations map lines to lines and preserve incidence.
$\square$
\vspace{5mm}

We must now investigate how $k$ transforms.  We have:\\

\vspace{5mm}
\noindent {\bf Lemma 5:} $\vect{N}'(\vect{x})=\frac{\matx{A}^{-T}\vect{N}_{\vect{x}}}{\sqrt{\vect{N}^{T}\matx{A}^{-1}\matx{A}^{-T}\vect{N}}}$\\

\vspace{5mm}
  \noindent {\bf Proof:}  Write $\Pi_{\vect{x}}$ for the tangent plane at $\vect{x}$.  For $\vect{u} \in \Pi_{\vect{x}}$, we have for non-unit normal $\vect{n}$,
  ${\vect{n}}(\vect{x})^T(\vect{u}-\vect{x})=0$. But $\vect{u}$ transforms to $\matx{A}\vect{u}+\vect{b}$, etc.
  and we must have $\vect{n}'(\vect{x})^T\left(\matx{A}(\vect{u}-\vect{x})\right)=0$, and the result follows by requiring the normal be a unit normal;
equivalently, normals are covariant. $\square$
\vspace{5mm}

  Write $\sigma_x$ for the largest singular value of $\matx{A}$,
  $\sigma_n$ for the smallest (which is positive), and  $c=\frac{\sigma_x}{\sigma_n}$ for the condition number.  We have

  \vspace{5mm}
  \noindent {\bf Lemma 6:} $(\sigma_n)\vect{N}^T \vect{d}_{\vect{u}\rightarrow\vect{v}}\leq
  \vect{N}'^T \vect{d}'_{\vect{u}\rightarrow\vect{v}}\leq ({\sigma_x})\vect{N}^T \vect{d}_{\vect{u}\rightarrow\vect{v}}$

  \vspace{5mm}
  \noindent {\bf Proof:}
  $ \vect{N}'^T \vect{d}'_{\vect{u}\rightarrow\vect{v}}=\frac{\vect{N}^T{\bf A}^{-1}{\bf A} \vect{d}_{\vect{u}\rightarrow\vect{v}}}{\sqrt{\vect{N}^T{\bf A}^{-1}{\bf A}^{-T}\vect{N}}}$ and the bound follows. $\square$
\vspace{5mm}
  
\noindent {\bf Lemma 7:} ${\sigma_n}^2 {\cal T}(dA_{\vect{y}})\leq dA'_{\vect{y}} \leq {\sigma_x}^2
{\cal T}(dA_{\vect{y}})$.

\vspace{5mm}
\noindent {\bf Proof:} $dA'_{\vect{y}}$ is the infinitesimal area spanned by two infinitesimal vectors $dx_1$ and $dx_2$.
The largest (resp. smallest) scaling of each length is by the largest (resp. smallest) singular value of $\matx{A}$.  $\square$
\vspace{5mm}

We then obtain the novel:

\vspace{5mm}
  \noindent {\bf Lemma 8:} {\em For affine transformations,  $c^{-4} {\bf K} ({\cal T}(f))\leq {\bf K}' ({\cal T}(f))\leq c^4 {\bf K} ({\cal T}(f)) $ for any $f_{\vect{s}}$}\\

  \vspace{5mm}
  \noindent {\bf Proof:} For any two vectors $\vect{u}'^T\vect{v}'=\vect{u}^T\matx{A}^T\matx{A}\vect{v}$.
  Then $\sigma^2_n \vect{u}^T\vect{v}\leq \vect{u}^T\matx{A}^T\matx{A}\vect{v}=\vect{u}'^T\vect{v}'\leq \sigma^2_x
\vect{u}^T\vect{v}$. Lemma 6 deals with the angles.  Write
\[
  {\bf K'}({\cal T}(f))=\int_{\cal D} k'(\vect{x}, \vect{y}) f(\vect{y}) d A'_{\vect{y}}
  \]
  where  \[
  k'(\vect{x}, \vect{y})=
 \frac{V'(\vect{x}, \vect{y})
    \left[\vect{N}'^T(\vect{x})\vect{d}'_{\vect{x}\rightarrow \vect{y}}\right]
    \left[\vect{N}'^T(\vect{y}) \vect{d}'_{\vect{y}\rightarrow \vect{x}}\right]
    }
    {
      \left[\vect{d}'^T_{\vect{x}\rightarrow \vect{y}}\vect{d}'_{\vect{x}\rightarrow \vect{y}}\right]^2
    }
\]
But
\begin{eqnarray*}
  \left[\begin{array}{c} \left[({\sigma_n}) \vect{N}^T(\vect{x}) \vect{d}_{\vect{x}\rightarrow \vect{y}}\right]\times\\
      \left[({\sigma_n}) \vect{N}^T(\vect{y}) \vect{d}_{\vect{y}\rightarrow \vect{x}}\right] \times \\
      \left[\sigma^2_n dA_{\vect{y}}\right]\end{array}\right] &\leq&
     \left[\begin{array}{c} \left[\vect{N}'^T(\vect{x})\vect{d}'_{\vect{x}\rightarrow \vect{y}}\right]\times\\
                                                                                                               \left[\vect{N}'^T(\vect{x})
                                                                                                                 \vect{d}'_{\vect{y}\rightarrow
                                                                                                                   \vect{x}}\right]\times\\
                                                                                                               dA'_{\vect{y}}\end{array}\right]\\
     &\leq&
\left[\begin{array}{c}  \left[({\sigma_x}) \vect{N}^T(\vect{x}) \vect{d}_{\vect{x}\rightarrow \vect{y}}\right]\times\\
    \left[({\sigma_x}) \vect{N}^T(\vect{y}) \vect{d}_{\vect{y}\rightarrow \vect{x}}\right] \times \\
    \left[\sigma^2_x dA_{\vect{y}}\right]\end{array}\right]
\end{eqnarray*}
       and
      \begin{eqnarray*}
\sigma_n^4              \left[\vect{d}^T_{\vect{x}\rightarrow \vect{y}}\vect{d}_{\vect{x}\rightarrow \vect{y}}\right]^2&\leq&        
                                                                                                                               \left[\vect{d}'^T_{\vect{x}\rightarrow \vect{y}}\vect{d}'_{\vect{x}\rightarrow \vect{y}}\right]^2\\
                                                                                                                        &=&
                                                                                                                   \left[\vect{d}^T_{\vect{x}\rightarrow \vect{y}}\matx{A}^{T}\matx{A} \vect{d}_{\vect{x}\rightarrow \vect{y}}\right]^2\\
        &\leq&\sigma_x^4              \left[\vect{d}^T_{\vect{x}\rightarrow \vect{y}}\vect{d}_{\vect{x}\rightarrow \vect{y}}\right]^2
      \end{eqnarray*}
      so that
      \begin{eqnarray*}
        c^{-4} {\cal T}({\bf K})({\cal T}(f))&=&
        \int_{\cal D}  \frac{\sigma_n^4}{\sigma_x^4} k(\vect{x}, \vect{y}) f(\vect{y})dA_{\vect{y}}\\
                                             &\leq&
\int_{\cal D}  k'(\vect{x}, \vect{y}) f(\vect{y})dA'_{\vect{y}}\\
&\leq&
\int_{\cal D}  \frac{\sigma_x^4}{\sigma_n^4} k(\vect{x}, \vect{y}) f(\vect{y})dA_{\vect{y}}\\
&=& c^4 {\cal T}({\bf K})({\cal T}(f))
\end{eqnarray*}                                                    
$\square$.
      \vspace{5mm}
      
which leads to

\vspace{5mm}
\noindent {\bf Lemma 9:} {\em With the notation above,
  \[
    \gennorm{B_{V}-B_{V'}} \leq \frac{(c^4-1)}{(1-p)^2}\gennorm{E}
  \]
}

\vspace{5mm}
\noindent {\bf Proof:}
\begin{eqnarray*}
  B_{V'}-B_{V}&=&\left({\cal T}(E)-E\right) + \left({\bf K}' B_{V'}-{\bf K} B_{V}\right)\\
  &=&{\bf K}' B_{V'}-{\bf K}'B_{V}+{\bf K}' B_{V}-{\bf K} B_{V}\\
  &&\mbox{(recall ${\cal T}(E)=E$)}\\
  &=&{\bf K}' (B_{V'} -B_{V})+{\bf K}' B_{V}-{\bf K} B_{V}
\end{eqnarray*}
so that
\[
  \left({\bf I}-{\bf K}'\right)\left(B_{V'}-B_{V}\right)=\left({\bf K}'-{\bf K}\right) B_{V}
  \]
  and
  \[
  \left(B_{V'}-B_{V}\right)={\bf M}'^{-1}\left({\bf K}'-{\bf K}\right) B_{V}.
  \]
    Notice that the kernel is everywhere non-negative, and for every point $\vect{x}$ and for any non-negative function $f$,  we have
    $0\leq ({\bf K}f) (x)$.  Since $E$ is non-negative, so is $B_{V}$.  From lemma 6, for non-negative $f$, we have
    $({\bf K}' f )(\vect{x})\leq c^4 ({\bf K} f)(\vect{x})$ at every point $\vect{x}$, so that
    ${\bf K}' f-{\bf K}f \leq (c^4-1){\bf K}f$ at every point.
    This means
\begin{eqnarray*}
  \gennorm{\left(B_{V'}-B_{V}\right)}
  &=& \gennorm{{\bf M}'^{-1}\left({\bf K}'-{\bf K}\right) B_{V}}\\
  &\leq& \frac{1}{(1-p')} \gennorm{\left({\bf K}'-{\bf K}\right) B_{V}}\\
  &\leq& \frac{1}{(1-p')} (c^4-1)\gennorm{B_{V}}\\
  &\leq& \frac{1}{(1-p')} (c^4-1)\frac{1}{(1-p)}\gennorm{E}\\
  &=& \frac{(c^4-1)}{(1-p)^2} \gennorm{E}
\end{eqnarray*}
where the last step works because $p'=p$, because $\rho'={\cal T}(\rho)=\rho$.
$\square$.
\vspace{5mm}

Finally, we can prove the entirely novel

\vspace{5mm}
{\bf Theorem 1:} {\em For $V=({\cal G}, \rho, E)$ and $V'=({\cal G}'({\cal G}, \matx{A}, \vect{b}), \rho', E')$, where
  $\epsilon_E\gennorm{E} = \gennorm{E-E'}$, $\epsilon_\rho=\mbox{sup}_{\cal D}\abs{\left[\rho-\rho'\right]}$, $p=\mbox{sup}_{\cal D} \rho$,
  $p'=\mbox{sup}_{\cal D}\rho'$, $c$ is the condition number of $\matx{A}$ (ratio of largest to smallest eigenvalues), 
  we have:}
\[
  \gennorm{B_{V}-B_{V'}}\leq\left[\begin{array}{c}
      \left[\frac{\epsilon_E}{1-p}\right]+ \\
      \left[\frac{\epsilon_\rho(1+\epsilon_E)}{(1-p)(1-p')}\right]+\\
     \frac{(c^4-1)(1+\epsilon_E)}{(1-p')^2}\end{array}\right]  \gennorm{E}
\]
\vspace{5mm}

\noindent {\bf Proof:} Write $V_1=({\cal G}, \rho, E')$, $V_2=({\cal G}, \rho', E')$. Then
\[
\gennorm{B_{V}-B_{V'}}\leq\left[\begin{array}{c}
    \gennorm{B_{V}-B_{V_1}}+\\
    \gennorm{B_{V_1}-B_{V_2}}+\\
    \gennorm{B_{V_2}-B_{V'}}\end{array}\right] \mbox{(triangle inequality)}
\]
and $\gennorm{E'}\leq (1+\epsilon_E)\gennorm{E}$
but
\begin{eqnarray*}
  \gennorm{B_{V}-B_{V_1}}&\leq& \frac{\epsilon_E}{1-p} \gennorm{E}\mbox{ (Lemma 1)}\\
  \\
  \gennorm{B_{V_1}-B_{V_2}}&\leq& \frac{\epsilon_\rho}{(1-p)(1-p')}\gennorm{E'} \mbox{ (Lemma 3)}\\
  &\leq& \frac{\epsilon_\rho}{(1-p)(1-p')}(1+\epsilon_E)\gennorm{E} \mbox{ (definitions)}\\
  \\
  \gennorm{B_{V_2}-B_{V'}}&\leq& \frac{(c^4-1)}{(1-p')^2}(1+\epsilon_E)\gennorm{E} \mbox{ (Lemma 9)}                          \end{eqnarray*}
$\square$
\vspace{5mm}

\section{Appendix III: Proof of Theorem 2}

We now establish that viewing shading fields for a set of similar scenes $V_i$ is enough to estimate a representation of the generators for a
given scene $V$.  It is important to have a representation of the probability that a scene will be illuminated in some particular way.
We use the general definition of a scene, above.  A scene $V$ is a tuple of geometry, albedo and luminaire model; write $V=\left({\cal G}, \rho,
{\cal E}\right)$, where ${\cal E}$ is the luminaire model.     The geometry ${\cal G}$ consists of a set
of surfaces parametrized in some way and a 2D domain for that parametrization; write ${\cal G}=(\vect{s}(\vect{x}), {\cal
  D})$.  The luminaire model captures the idea that lighting in a particular scene may change.  We have
${\cal E}=\left(E_1, \ldots E_{N_e},  P(\theta)\right)$ for a set of
potentially many basis luminaires and a probability distribution over the coefficients.
This information implies a probability distribution over luminaires for the scene 
$P(E)$.    Two scenes $V$ and $V'$ are similar if there is some affine transformation
${\cal A}, \vect{b}$ ( smallest singular value of ${\cal A}$ greater than zero) such that
${\cal G}'({\cal G}, \matx{A}, \vect{b})=(\matx{A} \vect{s}(\vect{x})+\vect{b})$ and if
${\cal E}={\cal E}'$.    Notice this second constraint does {\em not} require that the two scenes are lit in the same way, just that the
basis of luminaires is the same and the probability distribution over lighting coefficients is the same.  

We pass to a finite dimensional representation without loss of generality (to avoid having to deal with function spaces).
Choose some $N_o$ element orthonormal basis to represent the possible radiosity vectors for the scene (if $N_o=N_e$, the representation is exact,
because there is a finite dimensional space of luminaires).  In this basis, any particular radiosity $B(\vect{x})$ has
coefficient vector $\vect{b}$.  Write $\phi_i(\vect{x})$ for the $i$'th element of the basis; $\Phi=\left[\phi_1(\vect{x}), \ldots, \phi_{N_o}(\vect{x})\right]$; $\vect{b}_i$ for the coefficient vector representing $B_i(\vect{x})$ (which is the radiosity resulting from $E_i(\vect{x})$);
and $\matx{B}=\left[\vect{b}_1, \ldots, \vect{b}_{N_e}\right]$.

For any $r \leq N_e$, an {\bf effective generator matrix} for scene $V$ is the $N_o \times r$ matrix  $\matx{G}_{e, V, r}$  such that for any illumination condition producing radiosity represented by
$\vect{b}$ , there is some $\vect{w}$ such that the radiosity  represented by $\matx{G}_{e, V, r} \vect{w}$
is similar to the actual radiosity field.   In particular,
we want
\[
{\cal L}(\matx{G}_{e, V, r})=\expee{\mbox{inf}_{\vect{w}}\ltwonorm{\matx{G}_{e, V, r} \vect{w}-\matx{B}\theta}^2}{\theta \sim P(\theta)}
\]
to be small; $\matx{G}_{e, V, r}$ is clearly not unique.  We choose to have $\matx{G}_{e, V, r}$ orthonormal.

Assume we have $k$ radiosity fields  $B_i$ for  a scene under different, unknown, illumination conditions, for $k$ very large.  We assume that for the $i$'th field, $\theta_i \sim P(\theta)$ (which is not known).  We can clearly estimate an effective generator matrix by replacing the expectation with an average.  But we do not have $k$ images of the same scene.

\vspace{5mm}
{\bf Lemma 10:} {\em Using the notation above, write
  \[
  C_{\vect{b}}=\matx{B}\left(\expee{\theta \theta^T}{\theta \sim P(\theta)}\right) \matx{B}^T.\]
  Then:}
  \[
  \expee{\mbox{inf}_{\vect{w}}\ltwonorm{\matx{G}_{e, V, r} \vect{w}-\matx{B}\theta}^2}{\theta \sim P(\theta)}
  \]
  is equal to
  \[
\mbox{Tr}\left[C_{\vect{b}}\right] - \mbox{Tr}\left[\matx{G}_{e, V, r}^T C_{\vect{b}} \matx{G}_{e, V, r}\right]
\]

\vspace{5mm}
{\bf Proof:} Manipulation; note that
\[ C_{\vect{b}}=\expee{\matx{B}\theta \theta^T \matx{B}^T}{\theta \sim P(\theta)};\]
that
\[\expee{\theta^T\matx{B}^T\matx{B}\theta}{\theta \sim P(\theta)}=\mbox{Tr}\left[C_{\vect{b}}\right];\]
and that
\[\expee{\theta^T\matx{B}^T\matx{G}_{e, V, r}\matx{G}_{e, V, r}^T \matx{B}\theta}{\theta \sim P(\theta)}=\mbox{Tr}\left[\matx{G}_{e, V, r}^T C_{\vect{b}}\matx{G}_{e, V, r}\right].\]
$\square$
\vspace{5mm}

We cannot {\em know} ${\cal C}_{\vect{b}}$, but we do not need to.

{\bf Lemma 11:} {\em Using the notation above; writing $\vect{e}_i$ for the eigenvector of
  $C_{\vect{b}}$ corresponding to the $i$'th largest eigenvalue;
  and assuming that the eigenvalues of $C_{\vect{b}}$ are distinct, an optimal value
  of ${\cal L}(\matx{G}_{e, V, r})$ is given by}
\[
\left[\vect{e}_1, \ldots, \vect{e}_r\right] \matx{M}
\]
  {\em for any $r \times r$ matrix $\matx{M}$ of full rank such that $\matx{M}^T\matx{M}=I$}
  \vspace{5mm}
  
  {\bf Proof:} Follows directly from lemma 10.$\square$
  \vspace{5mm}

  {\bf Lemma 12:} {\em Using the notation above, and writing $\lambda_i$ for the $i$'th largest eigenvalue of
    $C_{\vect{b}}$,  the optimal value  of ${\cal L}(\matx{G}_{e, V, r})$ is given by}
\[
\sum_{i=r+1}^{N_e} \lambda_i
\]
\vspace{5mm}

  {\bf Proof:} Follows directly from lemma 10 and 11. $\square$
  \vspace{5mm}
  
  Assume we collect $N_s$ samples of radiosity $\hat{\vect{b}}=\matx{B}\theta$, where $\hat{\vect{b}}_{ i}$ is the $i$'th such sample, and $\theta \sim P(\theta)$.  These yield an estimate of $C_{\vect{b}}$, which can be accurate.

  \vspace{5mm}
  {\bf Lemma 13:} {\em  Write $\hat{C}_{\vect{b}, m}=(1/N_s) \sum_i \hat{\vect{b}}_i \hat{\vect{b}}_i^T$.  Then there is some constant $C_e$
    such that
  }
  \begin{eqnarray*}
    \expect{\hat{C}_{\vect{b}, m}-C_{\vect{b}, m}}&=&0 \\
    \mbox{ and } &&\\
    \expect{\frobnorm{\hat{C}_{\vect{b}, m}-C_{\vect{b}, m}}^2}&=&(1/N_s) C_e
  \end{eqnarray*}
  \vspace{5mm}
  
  {\bf Proof:} This follows from the weak law of large numbers
  $\square$
  \vspace{5mm}
  
  Now collect $N_s$ samples of radiosity from different scenes $V_i$, each one similar
  to the scene $V$, where $\matx{A}_i$ maps the geometry of $V$ to $V_i$.  Write $\hat{\vect{b}}_{i, V_i}$ for the $i$'th such sample.  Recall that similar scenes
  have the same basis of luminaires and the same distribution for $\theta$.
  These samples yield an estimate of $C_{\vect{b}}$, too.

  \vspace{5mm}
  {\bf Lemma 14:} {\em Given $\vect{b}$,  $\vect{c}$, $r$ and $\delta$ such that (a) $\ltwonorm{\vect{b}-\vect{c}}\leq \delta$ and
    (b) $\ltwonorm{\vect{b}}\leq r$, we have:}
  \[
  \frobnorm{\vect{b}\vect{b}^T-\vect{c}\vect{c}^T}\leq 3 \delta^2 r^2+4 \delta^3 r+\delta^4
  \]
  \vspace{5mm}
  
    {\bf Proof:} Write $\Delta=\vect{c}-\vect{b}$.  Then
    \begin{eqnarray*}
    \frobnorm{\vect{b}\vect{b}^T-\vect{c}\vect{c}^T}&=&\mbox{Tr}\left[\left(\vect{b}\vect{b}^T-\vect{c}\vect{c}^T\right)
      \left(\vect{b}\vect{b}^T-\vect{c}\vect{c}^T\right)^T\right]\\
    &=&\left(\begin{array}{c}(\vect{b}^T\vect{b})^2\\
      -2(\vect{b}^T\vect{b}+\Delta^T\vect{b})^2\\
      +
      (\vect{b}^T\vect{b}+\Delta^T\vect{b}+\Delta^T\Delta)^2
      \end{array}\right)\\
    &\leq& 4 \delta r^3 +3 \delta^2 r^2+2\delta^3 r+\delta^4
    \end{eqnarray*}
    $\square$.
    \vspace{5mm}
    
  {\bf Lemma 15:} {\em  Write $\hat{C}_{\vect{b}, m, s}=(1/N_s) \sum_i \hat{\vect{b}}_{i, V_i} \hat{\vect{b}}_{i, V_i}^T$.
    Using the notation above, and writing $p=\mbox{max}_i \left(\mbox{sup}_{\cal D} \rho_i\right)$;
    $\epsilon_\rho=\mbox{max}_i \left(\mbox{sup}_{\cal D}\abs{\left[\rho_V-\rho_{V_i}\right]}\right)$;
    $c$ for the largest condition number of $\matx{A}_i$.  Write
    \[
    \delta=\left(\frac{\epsilon_\rho+(c^4-1)}{(1-p)^2}\right) \mbox{ and } r=\frac{1}{(1-p)}
    \]
    Then
  }
  \[
  \frobnorm{\hat{C}_{\vect{b}, m, s}-C_{\vect{b}, m}}\leq N_o^2  \left( 4 \delta r^3 + 3 \delta^2 r^2 + 2 \delta^3 r+\delta^4\right)
  \]
  \vspace{5mm}
  
    {\bf Proof:} The $i$'th sample $\theta_i$ from $P(\theta)$ yields a luminaire $E$; write $B_{i, V}$ for the resulting radiosity in $V$ and $\vect{b}_{i, V}$ for its
    finite dimensional representation.  Theorem 1 yields a bound on $\gennorm{B_{i, V}-B_{i, V_i}}$, and so on $\gennorm{\vect{b}_{i, V}-\vect{b}_{i, V_i}}$.
    We have  $\gennorm{\vect{b}_{i, V}-\vect{b}_{i, V_i}}\leq \left(\frac{\epsilon_\rho+(c^4-1)}{(1-p)^2}\right)$.
    Furthermore, $\gennorm{\vect{b}_{i, V}}\leq 1/(1-p)$.   The result then follows from lemma 14 $\square$.
    \vspace{5mm}

    We can now compute an effective generator matrix $\hat{\matx{G}}$ from
    an estimate $\matx{C}_{\vect{b}, m, s}$ of $\matx{C}_{\vect{b}}$.  This estimate uses observations of
    multiple similar scenes.  The $\hat{\matx{G}}$ we obtain from data like this will not be the true $\matx{G}$, and we
    must determine what problems will result from using $\hat{\matx{G}}$ (which we can get) in place of $\matx{G}$ (which is the true best EGM).
    
    Theorem 2 establishes that using $\hat{\matx{G}}$ in place of $\matx{G}$ incurs bounded error.
Recall the expected error of using $\matx{G}$ as the effective generator matrix for estimate $\matx{C}$ is
      \[
        {\cal L}(\matx{G}; \matx{C})=\left(\mbox{Tr}\left[\matx{C}\right] - \mbox{Tr}\left[\matx{G}^T \matx{C} \matx{G}\right]\right)
        \]
        Now estimate the effective generator matrix using $\matx{C}_{\vect{b}, m, s}$ to obtain
        \[
        \hat{\matx{G}}=\begin{array}{c}\mbox{argmax}\\\matx{G}| \matx{G}^T\matx{G}=I
        \end{array} {\cal L}(\matx{G}; \matx{C}_{\vect{b}, m, s})
        \]
        Using $\hat{\matx{G}}$ to represent the radiosity of the scene will incur expected error
      \[
        {\cal L}(\hat{\matx{G}}; \matx{C}_{\vect{b}})=\left(\mbox{Tr}\left[\matx{C}_{\vect{b}}\right] - \mbox{Tr}\left[\hat{\matx{G}}^T \matx{C}_{\vect{b}} \hat{\matx{G}}\right]\right).
        \]
        
        \vspace{5mm}
        {\bf Theorem 2:} {\em  ${\cal L}(\hat{\matx{G}}; \matx{C}_{\vect{b}})-{\cal L}(\matx{G}; \matx{C}_{\vect{b}})$ is bounded.}
        \vspace{5mm}
        
        {\bf Proof:}

        We have
        \[
          {\cal L}(\hat{\matx{G}}; \matx{C}_{\vect{b}})-{\cal L}(\matx{G}; \matx{C}_{\vect{b}})=
          \mbox{Tr}\left[\hat{\matx{G}}^T \matx{C}_{\vect{b}} \hat{\matx{G}}\right]-
          \mbox{Tr}\left[\matx{{G}}^T \matx{C}_{\vect{b}} \matx{{G}}\right]
          \]

          Recall $\matx{G}$ is chosen to minimize ${\cal L}(\matx{G}; \matx{C}_{\vect{b}})$, and is orthornormal, so will consist of the
          eigenvectors of $\matx{C}_{\vect{b}}$ corresponding to the $r$ largest eigenvalues.   In turn,
          $\hat{\matx{G}}$ is chosen to minimize ${\cal L}(\hat{\matx{G}}; \matx{C}_{\vect{b}, m, s})$.  We must now find the
          $\hat{\matx{G}}$ so that $\mbox{Tr}\left[\matx{{G}}^T \matx{C}_{\vect{b}} \matx{{G}}\right]$ is minimized within the
          bound on $\matx{C}_{\vect{b}}-\matx{C}_{\vect{b}, m, s}$.  Any extremal $\hat{\matx{G}}$ must consist of eigenvectors of $\matx{C}_{\vect{b}}$
          (recall $\hat{\matx{G}}$ is orthornormal). 
          This yields an immediate but rather loose bound.
          Label the eigenvalues $\lambda_i$ of $\matx{C}_{\vect{b}}$ so that $\lambda_1\geq \lambda_2\geq \ldots \lambda_{N_{o}}$.
          then
          \[
          \dafinf{\hat{\matx{G}}| \hat{\matx{G}}^T\hat{\matx{G}}=\matx{I}} \mbox{Tr}\left[\matx{{G}}^T \matx{C}_{\vect{b}} \matx{{G}}\right]=\sum_{i={N_{o}}-r+1}^{{N_{o}}}\lambda_{i}
          \]
          so
          \[
            {\cal L}(\hat{\matx{G}}; \matx{C}_{\vect{b}})-{\cal L}(\matx{G}; \matx{C}_{\vect{b}})\leq
            \sum_{i=1}^{{N_{o}}}\lambda_i - \sum_{i={N_{o}}-r+1}^{{N_{o}}}\lambda_{i}.
            \]
            $\square$
            
This bound is quite loose, but for the purposes of this paper, the existence of a bound is enough. Some quite interesting mathematics
is possible in tightening the bound, and there may be practical consequences.

Notice that, writing $\Phi(\matx{G}; \matx{C})=\mbox{Tr}\left[\matx{C}\right]-\mbox{Tr}\left[\matx{G}^T\matx{C}\matx{G}\right]$, with
$\matx{C}$ symmetric, diagonalized as $\matx{U}^T\Lambda \matx{U}$; then $\Phi(\matx{U}\matx{G}; \Lambda)=\Phi(\matx{G}; \matx{C})$.  This means we can work with diagonal $\matx{C}$ WLOG.  In turn,  determining the value of
${\cal L}(\hat{\matx{G}}; \matx{C}_{\vect{b}})-{\cal L}(\matx{G}; \matx{C}_{\vect{b}})$ requires thinking about shuffling eigenvalues.
Any bound will follow from reasoning about permutations in the order of eigenvectors achieved
          by the error, and the estimation error may not be large enough to permute the eigenvalues so aggressively.
             A more interesting and tighter bound can be obtained as
            \[
            \sum_{i=1}^{{N_{o}}}\lambda_i - v
            \]
            where $v$ is the value of a linear program representing the permuation of the
            eigenvalues.  Write $\vect{1}_r$ for the $N_o$ dimensional vector consisting of $r$ ones and $N_o-r$ zeros in order,
            $\lambda=\left[\lambda_1, \ldots, \lambda_{N_o}\right]$ (eigenvalues of $\matx{C}_{\vect{b}}$), $\matx{P}$ for a permutation matrix,
            $\delta$ for a vector of offsets, $D$ for $\frobnorm{\matx{C}_{\vect{b}}-\matx{C}_{\vect{b}, m, s}}^2$, then $v$ is the value of:
            \begin{eqnarray*}
              \dafmin{\matx{P}, \delta, \vect{m}} \vect{1}_r^T \matx{P} \lambda\\
              \mbox{subject to}\\
              \sum_i p_{ij}&=&1\\
              \sum_j p_{ij}&=&1\\
              p_{ij}&\geq &0\\
              \vect{u}&=&\matx{P}\lambda\\
              \vect{m}&=&\vect{u}+\delta\\
              m_1\geq m_2 &\geq& \ldots m_{N_o}\\
              \sum_i \delta_i^2 \leq D.
            \end{eqnarray*}
            This problem   finds the permutation of eigenvalues that produces the smallest value within the budget $D$.
            For $D\geq 0$, the program is feasible ($\matx{P}=\matx{I}$ is always a feasible point), and Birkhoff's theorem guarantees that a there is a solution with integer $\matx{P}$.  Interestingly, for sufficiently small $D$,
            $v=\sum_{i=1}^{{N_{o}}}\lambda_i$, meaning the error bound is zero! This occurs when $D$ is smaller than $\lambda_{N_o}-\lambda_{N_o+1}$.
            The quantized character of this bound means that quite small improvements in estimation of $\matx{C}_{\vect{b}, m, s}$ could result in large
            improvements in the practical behavior of the method.   

\section{Appendix IV: The Local FID}

\textbf{Definition:}  Assume we obtain a collection of images ${\cal R}$ from
a ground truth collection ${\cal O}$ by relighting a single image $I_k$ and compute $\fid({\cal O}, {\cal R})$.  This value -- which is
non-negative, and is ideally zero -- measures the extent to which the relighted image is realistic.  Furthermore,
different relightings of the same image will have different values. Write $\mu_O$ for the mean and $C_O$ for the covariance of ${\cal O}$ in the chosen embedding, etc. 
We have
\begin{eqnarray*}
  \fid({\cal O}, {\cal R})&=&\ltwonormsq{\mu_R-\mu_O} +
  \\&&\trace{C_O + C_R - 2 \msqrt{C_O C_R}}.
\end{eqnarray*}

\textbf{Computing difficulties:} Numerical problems occur because $C_R$ is very similar to $C_O$,  and $C_O$ has many small eigenvalues; this means
there are many eigenvalues of $C_OC_R$ that are very close to zero, which is a reliable source of trouble for matrix square root methods.  One typically gets complex estimates.

\textbf{Series estimate:}  Write $\vect{x}(I_k)$ for the embedding of $I_k$, $R_k$ for the result of relighting $I_k$, and
$\vect{d}_k=\vect{x}(R_k)-\vect{x}(I_k)$.  Then $\mu_R-\mu_O=\frac{\vect{d}_k}{N}$, and
\begin{eqnarray*}
  C_R&=&C_O\\
  &&+\frac{1}{N}\left(\begin{array}{c}\vect{d}_k (\vect{x}(I_k)-\mu_0)^T+\\ (\vect{x}(I_k)-\mu_0)\vect{d}_k^T+\\
    \vect{d}_k \vect{d}_k^T\end{array}\right)\\
  &&-\frac{1}{N^2}\vect{d}_k \vect{d}_k^T\\
  &=&C_O+\frac{1}{N} U +\frac{1}{N^2} V
\end{eqnarray*}
We seek an approximation to $\msqrt{C_O C_R}$ of the form $C_O+\frac{1}{N} M_1 +\frac{1}{N^2} M_2 +\ldots$.  By squaring
and matching terms, we have
\[
C_O M_1 +  M_1 C_O=C_O U
\]
and
\[
C_O M_2 +  M_2 C_O+M_1 M_1=C_O V \
\]
so $2 \trace{M_1}=\trace{U}$ and first order terms cancel.  Similarly, $\trace{V}=2
\trace{M_2}+\trace{C_O^{-1}M_1^2}$, and we have
\[
  \fid({\cal O}, {\cal R})=\frac{1}{N^2}(\vect{d}_k^T\vect{d}_k + \trace{C_O^{-1}M_1^2}+ O({\frac{1}{N}}^3)
\]
and we define the local FID at the $k$'th image as
\[
  \lfid{{\cal O}}(I_k)=(\vect{d}_k^T\vect{d}_k + \trace{C_O^{-1}M_1^2}).
  \]
  But we must evaluate $M_1$.  We have $C_O M_1 +  M_1 C_O=C_O U$,  which is a symmetric Sylvester equation.  While this is a linear system
  in $M_1$, it is inconveniently large.  But, for $R$ a rotation and $G$ general, $\msqrt{R G R^T}=R\msqrt{G}R^T$, so
  \[
  \trace{RC_OR^T + RC_RR^T-2 \msqrt{R C_O R^T R C_R R^T}}
  \]
  is equal to
  \[
  \trace{C_O + C_R-2 \msqrt{C_O C_R }}
  \]
  and the FID is not affected by a rotation in the embedding space, so we can assume without loss of generality that $C_O$ is diagonal.  In
this case, we can compute $M_1$; write $a_{ij}$ for the $i$, $j$'th component of $A$ etc., and we have $ m_{ij}=c_{ii}
u_{ij}/(c_{ii}+c_{jj})$, and computing the local FID at a particular image is relatively straightforward.

{\bf LFID is a good predictor of human preference:}  We show human subjects a pair of
images (A, B) {\em of the same scene}, and record which is regarded as more realistic.  Images are chosen uniformly at
random from the original, and four relightings.  We have 5000 runs of 10 choices, with anonymous subjects.  The scatter
plot shows frequency that A is preferred against LFID(A)-LFID(B) for 500 quantiles of the difference (so frequencies are
estimated with 100 points).  Subjects reliably choose the image with lower LFID, though probabilities at large
differences saturate.  The variation in frequencies suggest that other predictors might be required to score realism of
an image transformation more accurately.  

\section{Appendix V: The Gloss Term}

Experimental results suggest strongly our method can remove or enhance glossy effects.  Theorem 1 covers global illumination in a diffuse world.
The methods used to prove this theorem extend to cover single-bounce radiance from a luminaire for some BRDF models, including natural gloss models,
if all BRDF's have the property
  \[
    \left[(\omega' \cdot \omega)(\overline{\omega'} \cdot \overline{\omega})\right]^\alpha
    \leq \frac{b(\vect{x}, \omega', \overline{\omega}')}{b(\vect{x}, \omega, \overline{\omega})}\leq \left[\frac{1}{(\omega' \cdot \omega)(\overline{\omega'} \cdot \overline{\omega})}\right]^{\alpha}.
    \]
Notice all diffuse surfaces have this property for any finite $\alpha\geq 0$.
\bibliographystyle{IEEEtran}
\bibliography{bigdafmerge}
\end{document}